%% file: main.tex
\documentclass[letterpaper]{article} 
\usepackage{lineno}
\usepackage{aaai2026}  
\usepackage{times}  
\usepackage{helvet}  
\usepackage{courier}  
\usepackage[hyphens]{url}  
\usepackage{graphicx} 
\usepackage{natbib}  
\usepackage{caption} 
\usepackage{algorithm}
\usepackage{algorithmic}
\usepackage{pifont}
\usepackage{rotating}
\usepackage{amsfonts,amssymb}
\usepackage{amsmath}
\usepackage{color}
\usepackage[table]{xcolor}
\usepackage{booktabs}
\usepackage{multirow}
\usepackage{multicol}
\usepackage{makecell}
\usepackage{rotating}
\usepackage{colortbl}
\usepackage{times}  
\usepackage{helvet}  
\usepackage{courier}  
\usepackage[hyphens]{url}  
\usepackage{graphicx} 

\usepackage{pifont}

\newcommand{\cc}{\color[rgb]{0,0.6,0.3}\checkmark}
\newcommand{\xx}{\color[rgb]{0.6,0,0}{\ding{55}}}
\definecolor{mygray}{gray}{.6}
\definecolor{myblue}{RGB}{89,158,254}
\definecolor{mygreen1}{RGB}{81,150,111}
\definecolor{mygreen2}{RGB}{93,174,86}
\definecolor{myred}{RGB}{160,0,0}
\definecolor{myyellow}{RGB}{227,207,87}
\usepackage{newfloat}
\usepackage{listings}
\DeclareCaptionStyle{ruled}{labelfont=normalfont,labelsep=colon,strut=off} 
\floatstyle{ruled}
\newfloat{listing}{tb}{lst}{}
\floatname{listing}{Listing}
\title{CompTrack: Information Bottleneck‑Guided Low-Rank \\Dynamic Token Compression for Point Cloud Tracking}

\author{
    Sifan Zhou\textsuperscript{\rm 1,2},
    Yichao Cao\textsuperscript{\rm 3},
    Jiahao Nie\textsuperscript{\rm 4},
    Yuqian Fu\textsuperscript{\rm 5},
    Ziyu Zhao\textsuperscript{\rm 1,2},
    Xiaobo Lu\textsuperscript{\rm 1,2}\thanks{Corresponding author: Xiaobo Lu.}, 
    Shuo Wang\textsuperscript{\rm 6}
}

\affiliations{
    \textsuperscript{\rm 1}School of Automation, Southeast University
    \textsuperscript{\rm 2} Key Laboratory of Measurement and Control of Complex Systems of Engineering, Ministry of Education, Southeast University, Nanjing, China\\
    \textsuperscript{\rm 3}Central South University 
    \textsuperscript{\rm 4}Zhejiang University of Finance and Economics \\
    \textsuperscript{\rm 5}INSAIT, Sofia University "St. Kliment Ohridski"
    \textsuperscript{\rm 6}Mininglamp Technology \\
    \ sifanjay@gmail.com, xblu2013@126.com, wangshuo.e@mininglamp.com}

\usepackage{bibentry}

\begin{document}

\maketitle

\input{sec/0_abstract}

\input{sec/1_intro} 
\input{sec/2_related_work}

\input{sec/3_method}

\input{sec/4_experiments}
\input{sec/5_conclusion}

\bibliography{aaai2026}
\input{sec/6_supp}

\end{document}

%% file: sec/0_abstract.tex
\begin{abstract}
3D single object tracking (SOT) in LiDAR point clouds is a critical task in for computer vision and autonomous driving. Despite great success having been achieved, the inherent sparsity of point clouds introduces a dual-redundancy challenge that limits existing trackers: \textbf{(1)} vast spatial redundancy from background noise impairs accuracy, and \textbf{(2)} informational redundancy within the foreground hinders efficiency. To tackle these issues, we propose \textbf{CompTrack}, an novel end-to-end framework that systematically eliminates both forms of redundancy in point clouds. First, CompTrack incorporates a \textit{Spatial Foreground Predictor (SFP)} module to filter out irrelevant background noise based on information entropy, addressing spatial redundancy. Subsequently, its core is an \textit{Information Bottleneck-guided Dynamic Token Compression (IB-DTC)} module that eliminates the informational redundancy within the foreground. Theoretically grounded in low-rank approximation, this module leverages an online SVD analysis to adaptively compress the redundant foreground into a compact and highly informative set of proxy tokens.Extensive experiments on KITTI, nuScenes and Waymo datasets demonstrate that CompTrack achieves top-performing tracking performance with superior efficiency, running at a real-time 90 FPS on a single RTX 3090 GPU.

\end{abstract}

%% file: sec/1_intro.tex
\vspace{-3mm}
\section{Introduction}
\label{sec:intro}
Visual perception has wide application in computer vision~\cite{javed2022visual, brodermann2025cafuser,zhang2023multi,tao2023dudb,liao2025convex,dai2025unbiased,yin2025knowledge}, artificial intelligence~\cite{zhou2025neural,zhou2025opening,lu2023tf,lu2024mace,Qiu2025IntentVCNet,Wang2025OneImage} and robotics~\cite{Wang_Zhang_Dodgson_2025,ScanTD,jmse13091615,In-Pipe,ZHANG2025130845,TCP,zhao2024balf,zhao2023benchmark}. Among them, Single object tracking (SOT) based on LiDAR point clouds is a fundamental task in 3D computer vision, with broad applications in autonomous driving and robotics~\cite{cui2019point,ptt,fan2025beyond,focustrack,liudifflow3d,mamba4d,zhou2023fastpillars}. Given the annotated target in the first frame of a point cloud sequence, the goal of SOT is to continuously localize the same object in subsequent frames~\cite{m2track,m2track++, p2p}. Despite recent progress, accurately and efficiently tracking objects in sparse point clouds remains a key challenge, stemming from unaddressed data redundancy.

\begin{figure}[t]
\centering
\includegraphics[width=0.98\linewidth]{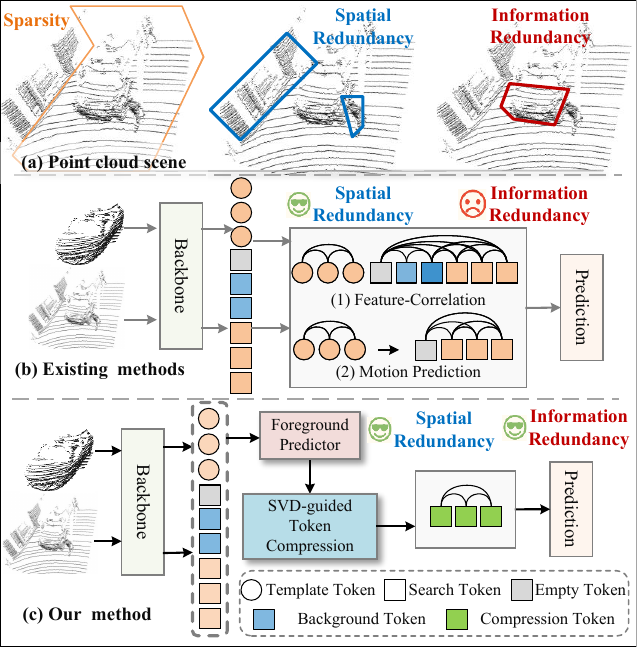}
\caption{
\textbf{(a)} The inherent sparsity of point clouds introduces dual challenges: spatial redundancy from irrelevant background and informational redundancy from repetitive geometries in foreground. \textbf{(b)} Existing SOT methods overlook the information redundancy, which limits their efficiency. \textbf{(c)} Our proposed \textbf{CompTrack} framework tackles both spatial and informational redundancy.}

\vspace{-20pt}
\label{fig1}
\end{figure}

Early approaches for 3D SOT primarily rely on point-based representations. As a pioneering method, SC3D~\cite{sc3d} introduces appearance matching to the 3D SOT by identifying the most similar region in subsequent frames based on the given template. Based on this, P2B~\cite{p2b} advances appearance matching by integrating a region proposal network (RPN)~\cite{votenet}  to generate high-quality candidates. Due to its strong performance, a wide range of follow-up works are proposed, such as PTT~\cite{ptt} and MBPTrack~\cite{mbptrack}. Recognizing the limitations of appearance-matching, which can be unreliable with sparse or ambiguous features,, motion-centric paradigm~\cite{m2track, m2track++,p2p}  emerged. These methods instead formulate the tracking as the task of modeling of target's motion between consecutive frames.

However, the methods above largely ignore two distinct forms of redundancy that arise from the inherent sparsity of LiDAR data. As shown in Fig.~\ref{fig1} (a), this challenge manifests on two distinct levels. \textbf{(1) Spatial redundancy},  caused by the vast number of irrelevant background and empty points that inundate the few features representing the actual target. This creates a severe signal-to-noise problem and leads to computationally wasteful processing. \textbf{(2) Informational redundancy}, which stems from the fact that not all points on an object are equally informative within the foreground points. Consider a vehicle (Fig.~\ref{fig1} (a))): points on a large, flat surface like a hood offer ambiguous localization cues, as the local geometry remains unchanged under translation. In contrast, points on corners or edges—where multiple surfaces intersect—provide unique structural information and serve as a reliable indicator of the object location. This is similar to the \textit{aperture problem} in optical flow and has long been pointed out in 2D image recognition~\cite{lowe1987three,harris1988combined}. Consequently, the foreground representation is dominated by a large number of these less informative, highly correlated points from simple surfaces. This results in significant informational redundancy and a low-rank feature structure. In turn, existing methods (Fig.~\ref{fig1} (b)) primarily address only spatial redundancy, inherently leaving the challenge of informational redundancy unsolved. 

To tackle this dual challenge, we propose \textbf{CompTrack}, a novel framework designed to eliminate both the spatial and information redundancy. As show in Fig.~\ref{fig1} (c),  CompTrack first employs a \textit{Spatial Foreground Predictor (SPF)} to address spatial redundancy by filtering the vast amount of background noise. Secondly, its core is an \textit{Information Bottleneck-guided Dynamic Token Compression (IB-DTC)} module that resolves informational redundancy. Grounded in the principle of Information Bottleneck and optimal low-rank approximation, this module adaptively compresses the redundant foreground into a compact and highly informative set of proxy tokens for precise and efficient tracking. By systematically removing both forms of redundancy, CompTrack achieves an exceptional balance between performance and latency.  Our CompTrack establishes new state-of-the-art performance on the large-scale nuScenes~\cite{nuScenes} and Waymo~\cite{waymo} benchmarks, while achieving highly competitive results on the KITTI~\cite{kitti}. Meanwhile, it also runs at a high real-time speed of 90 FPS on a single RTX 3090 GPU, which is 1.4X faster than previous leading methods~\cite{p2p}. The main contributions can be summarized as follows:

\begin{enumerate}
    \item We propose \textbf{CompTrack}, a novel end-to-end framework that, for the first time, tackles the dual challenges of spatial and information redundancy inherent to sparse point clouds during object tracking.
     \item We design a \textbf{Spatial Foreground Predictor} that effectively eliminates spatial redundancy. It was trained with a center-guided Gaussian circle, and justified by information entropy analysis showing noise filtering.

    \item We introduce a theoretical \textbf{Information Bottleneck-guided Dynamic Token Compression} module that is theoretically grounded in the Information Bottleneck principle. It uniquely synergizes an online SVD analysis to dynamically determine the optimal compression ratio with a learnable query mechanism that performs task-specific adaptation, resolving informational redundancy.
    
    \item Extensive experiments on three popular benchmarks demonstrate that CompTrack achieves leading performance, operating at high 90 FPS and demonstrating a superior trade-off between accuracy and speed.
\end{enumerate}

%% file: sec/2_related_work.tex
\begin{figure*}[t]
\centering
\includegraphics[width=0.9\linewidth]{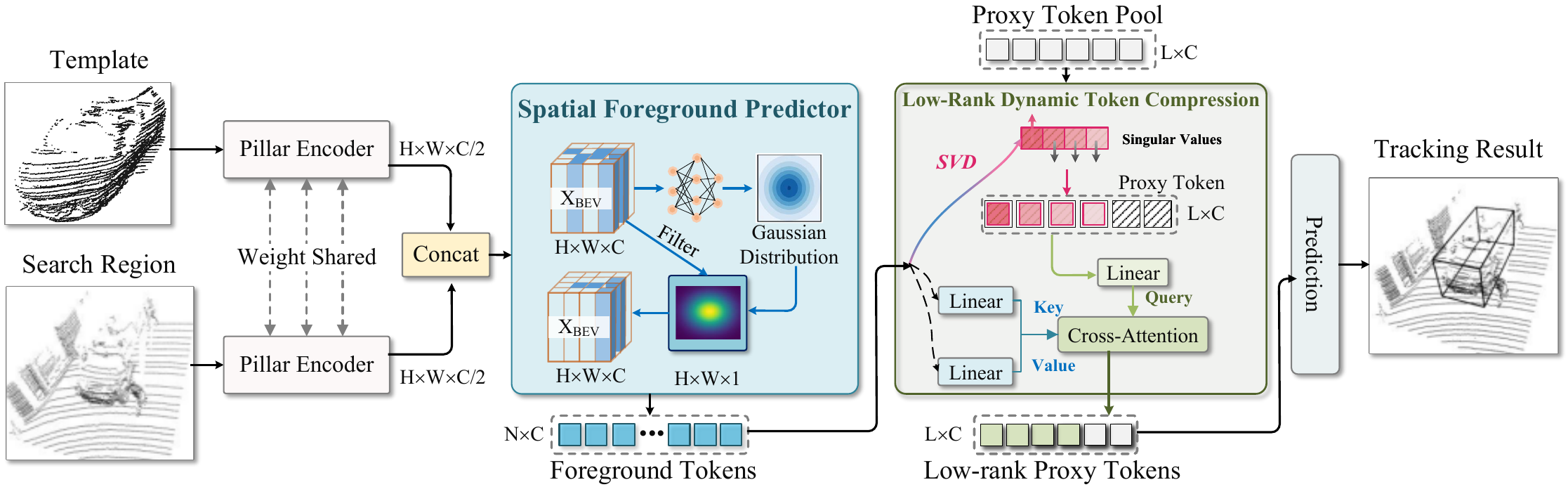}
\caption{\textbf{Overall architecture of our proposed CompTrack.} It consists of two main designs: (1)\textit{Spatial Foreground Predictor (SFP)} that filters irrelevant background to decrease the spatial redundancy, and (2) \textit{Information Bottleneck-guided Low-rank Dynamic Token Compression (IB-DTC)} module that compresses the foreground into a more compact, low-rank representation.
}
\label{framework}
\vspace{-3mm}
\end{figure*}

\vspace{-4mm}
\section{Related Work}
\paragraph{3D Point Cloud Object Tracking.} Early methods in 3D SOT primarily adopt the appearance matching paradigm, where the goal is to compare features of the template and candidates to identify the most representative instance. As a pioneer, SC3D~\cite{sc3d} introduces the first Siamese architecture in 3D tracking. P2B~\cite{p2b} and 3D-SiamRPN~\cite{3dsiamrpn} leverage Region Proposal Network (RPN) to generate high-quality 3D proposals in an end-to-end manner. Those inspired a wave of follow-up works~\cite{osp2b,cmt,hu2025mvctrack,pillartrack,synctrack}. For instance, BAT~\cite{bat} augmented correlation learning by encoding the size priors, while methods like PTT~\cite{ptt,ptt-journal}, and STNet~\cite{stnet}, GLT-T~\cite{glt} explored various attention mechanisms to improve feature representation. Recent works such as CXTrack~\cite{cxtrack} and MBPTrack~\cite{mbptrack} emphasized the importance of context and memory, respectively. Meanwhile, M$^2$Track~\cite{m2track,m2track++} series proposes motion-centric frameworks that track objects without relying on appearance similarity. Instead, they predict object future positions based on motion priors, which is particularly beneficial in scenarios where appearance information is ambiguous. Despite their progress, existing appearance- and motion-based trackers neglect the inherent sparsity of point clouds, leading to struggle with background noise and foreground redundancy.

\paragraph{Efficient Visual Tracking.} The pursuit of efficiency is a significant trend in the broader Tracking domain, driven by the demand for high-performance, low-latency tracking on resource-constrained devices. In 2D SOT, a new wave of efficient algorithms~\cite{blatter2023efficient,zhu2025exploring,tan2025xtrack} has emerged to address this. Early approaches such as ECO~\cite{eco} and ATOM~\cite{atom} demonstrated real-time tracking capabilities on edge devices, but fell short in terms of accuracy compared to recent state-of-the-art methods. Recently, efficient trackers such as LightTrack~\cite{lighttrack}, which leverages neural architecture search (NAS), and FEAR~\cite{fear}, which introduces a dual-template representation and pixel-wise fusion block, have developed betweer trade-off between speed and accuracy. HiT~\cite{hit} proposes a bridge module to integrate high-level semantics with fine-grained details, utilizing large-stride down-sampling backbones in efficient tracking. LoRAT~\cite{LoRAT} leverages LoRA~\cite{hu2022lora} to fine-tune partial parameters without extra latency. While the 2D domain has made significant strides, efficiency has been a less explored frontier in 3D SOT. We argue that the inherent sparsity of point clouds presents a unique opportunity to bridge this gap. As a result, we propose the CompTrack, enabling a better trade-off between efficiency and accuracy on 3D tracking.

%% file: sec/3_method.tex
\vspace{-2mm}
\section{Methodology}

\subsection{CompTrack Architecture} 
Given a template point cloud $\mathcal{P}^{t}\in \mathbb{R}^{N_{t}\times 3}$ and its corresponding 3D bounding box (BBox) $B_t = (x_t,y_t,z_t,w_t,h_t,l_t,\theta_t)$ in the first frame, where $(x_t,y_t,z_t)$ and $(w_t,h_t,l_t)$ denote the center coordinate and size, and $\theta_t$ is the rotation angle. LiDAR-based 3D single object tracking (3D SOT) aims to locate the object within the search region $\mathcal{P}^{s}\in\mathbb{R}^{N_{s}\times 3}$ and output a 3D BBox $B_s$ frame by frame. Note that for the consistent size of the given target in all frames, we can output only 4 parameters to represent $B_s$. Our propose \textbf{CompTrack} is designed to address this task by tackling the dual challenges of \textbf{spatial} and \textbf{informational redundancy} inherent to sparse LiDAR data. As depicted in Fig~\ref{framework}, CompTrack is composed of two main stages: (1)\textit{Spatial Foreground Predictor (SFP)} that filters irrelevant background to decrease the spatial redundancy, and (2) \textit{Information Bottleneck-guided Dynamic Token Compression (IB-DTC)} module that compresses the key foreground into a more compact, high-rank representation.

\noindent\textbf{Pillar Encoder.}
We transform the raw point clouds into Bird's-Eye-View (BEV) feature maps for computational efficiency. Specifically, we follow PillarHist~\cite{zhou2025pillarhist} to process irregular points due to its preservation of fine-grained geometries and efficiency. The input points $\mathcal{P} \in \mathbb{R}^{N \times 3}$ are scattered to form a 2D pseudo-image as a BEV feature. This approach circumvents the need for computationally expensive 3D convolutions used in voxel-based methods~\cite{voxeltrack}. The resulting BEV features for the template and search region are denoted as $F_{t}$ and $F_{s}\in \mathbb{R}^{H \times W \times C}$. Details can refer to ~\cite{zhou2025pillarhist}.

\subsection{Spatial Foreground Predictor (SFP)}
\label{sec:method-SFP}
\begin{figure}[h]
\centering
\includegraphics[width=0.9\linewidth]{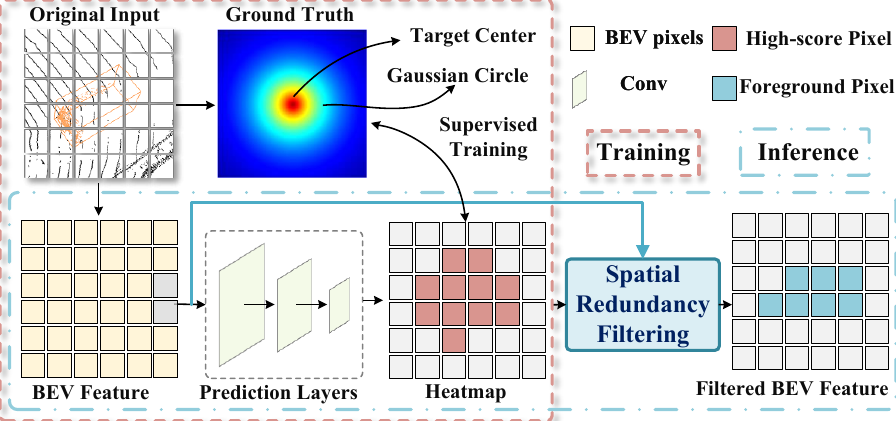}
\caption{Illustration of the spatial foreground predictor (SFP). SPF removes the spatial redundancy by filtering irrelevant background information.}
\label{fig:SPF}
\vspace{-1mm}
\end{figure}

As discussed in Sec. 1 and illustrated in Fig.~\ref{fig:SPF}, the inherent sparsity of point clouds leads to significant spatial redundancy in the BEV representation. This can be formally understood from an information-theoretic perspective. Given a low occupancy probability $p \ll 1$ for any BEV pillar, the total entropy (information content) of the BEV map is:
\vspace{-1mm}
\begin{align}
    H(\mathbf{X}) = HW\bigl[H_b(p)+p\,H_{\mathrm{fg}}\bigr].
    \label{eq:bev-entropy}
\end{align}
where $H_b(p)$ is the binary entropy of occupancy and $H_{\mathrm{fg}}$ is the entropy of foreground. For small $p$, empty or background pillars contribute negligible information. Consequently, filtering these locations is theoretically \textbf{information-lossless} and yields a cleaner feature for tracking.

To practically realize this information-theoretic filtering, we propose the \textbf{Spatial Foreground Predictor (SFP)}, a lightweight module designed to generate a spatial attention map that assigns an importance score to each location in the BEV grid. The SFP, denoted as $\mathbf{F}_{pred}$ with parameters $\mathbf{\Theta}_{pred}$, is a lightweight CNN with group convolutions~\cite{zhang2018shufflenet}, which results in a negligible computational overhead. SFP module takes the concatenated BEV features $\mathbf{X}_{BEV}=\text{cat}(\mathbf{F}_{t}, \mathbf{F}_{s})$ as input and outputs a spatial-wise importance heatmap $\mathbf{Y}_{\text{pred}} \in [0,1]^{H \times W}$. The refined search feature map, $\hat{\mathbf{F}}_{s}$, is then obtained via element-wise modulation:
\vspace{-1mm}
\begin{align}
  \mathbf{Y}_{pred} &= \mathbf{F}_{pred}(\mathbf{X}_{BEV}; \Theta_{pred}), \\
    \hat{\mathbf{F}}_{s} &= \mathbf{F}_s \odot \mathbf{Y}_{pred}.
  \label{equ:predictor_infer}
\end{align}

\begin{figure*}[t]
\centering
\includegraphics[width=0.9\linewidth]{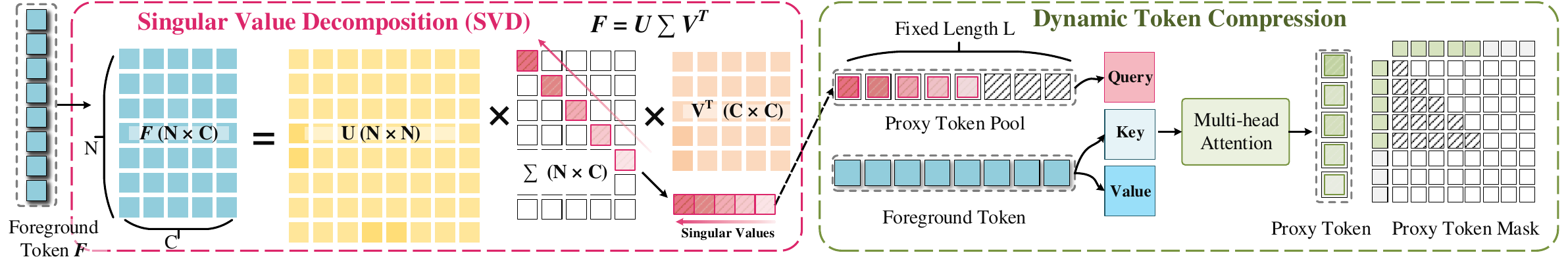}
\caption{Illustration of the proposed information bottleneck-guided dynamic token compression.}
\label{fig3}
\vspace{-3mm}
\end{figure*}

This operation dynamically enhances target-relevant foreground features while suppressing background noise with minimal computational overhead.

\paragraph{SFP Training.} 
As discussed in~\cite{zhao2023ada3d}, the center of the bounding box is of high importance and the importance spreads to the local region. Naturally, inspired by CenterPoint~\cite{center}, we supervise the predictor's training by generating a ground-truth heatmap $\mathbf{M}_{gt}$. Instead of a binary mask, $\mathbf{M}_{gt}$is formed by rendering a 2D Gaussian circle with a peak centered at each ground-truth box center $(u,v)_{i}$, which can be formulated as:
\vspace{-3mm}
\begin{align}
  \begin{split}
  \mathbf{M}_{gt}=\sum_{b} \mathcal{G}((u,v),\sigma), \\
  \end{split}
  \label{equ:predictor_train}
\end{align}
where $b$ is the ground-truth bounding box, and $\mathcal{G}$ is the 2D gaussian function with radius $\sigma$ adaptively determined by the box size. The predictor is then trained by minimizing the Mean Squared Error (MSE) loss between the predicted heatmap and the ground-truth: $\mathbf{L}_{pred} = \text{MSE}(\mathbf{Y}_{pred}, \mathbf{M}_{gt})$.

\subsection{Information Bottleneck-guided Dynamic Token Compression (IB-DTC)}
Having filtered the \textit{spatial redundancy}, we next tackle the more challenging of \textit{informational redundancy} within the foreground. As discussed in sec \ref{sec:intro}, not all foreground points are equally informative for localization. Points on geometrically simple surfaces (e.g., a vehicle's hood) provide ambiguous cues, whereas points at structural intersections (e.g., corners) are highly informative. This disparity leads to a foreground feature matrix, $\mathbf{X}_{fg}$, that is redundant and possesses a \textit{low effective rank}. To this end, we propose the IB-DTC module. Its central goal is to dynamically compress the sparse yet redundant foreground sequence, denoted as $\mathbf{X}_{fg} \in \mathbb{R}^{N \times C}$, into a compact, information-dense sequence $\mathbf{X}_{proxy} \in \mathbb{R}^{K \times C}$ (where $K \ll N$),  while preserving the most discriminative characteristics essential for robust tracking. This process is implemented via a learnable, end-to-end differentiable mechanism.

\subsubsection{Information Bottleneck Objective and Low-Rank Approximation.} 

For the foreground token set $\mathbf{X}_{fg}$, our compression objective aligns naturally with the \emph{Information Bottleneck} (IB) principle, which seeks a representation $\mathbf{X}_{comp}$:
\vspace{-3mm}
\begin{align}
\min_{g} ; I(\mathbf{X}_{fg}; \mathbf{X}_{proxy}) \quad \text{s.t.} \quad I(\mathbf{X}_{proxy}; \mathbf{y}) \geq I_0,
\label{eq:ib_objective}
\end{align}

\noindent where $\mathbf{y}$ is the network output, denotes the target motion state to be predicted (e.g., center offset, orientation). The \textit{IB} principle directs us to discard all information in $\mathbf{X}_{fg}$ that is not predictive of y. In other words, we seek to retain only the subset of information in $\mathbf{X}_{fg}$ that is predictive of the state $\mathbf{y}$, and discard irrelevant or redundant components. 

Directly optimizing the \textit{IB} objective is intractable. We therefore employ a practical and powerful surrogate: optimal low-rank approximation. The key insight is that the informational redundancy in $\mathbf{X}_{fg}$ corresponds to the low-variance components in its feature space. According to the Eckart-Young Theorem~\cite{eckart1936approximation}, the optimal rank-$K$ approximation of $\mathbf{X}_{fg}$, denoted as $\mathbf{X}_{proxy}$ is found by truncating its Singular Value Decomposition (SVD)~\cite{golub2013matrix} to its top K components:
\vspace{-1mm}
\begin{align}
\mathbf{X}_{proxy} = U \Sigma V^T,
\end{align}
where $U \in \mathbb{R}^{N \times N}$, $\Sigma \in \mathbb{R}^{N \times C}$ (containing the top $K$ largest singular values), and $V \in \mathbb{R}^{C \times C}$ are the truncated versions of $U, \Sigma, V$, respectively. The theorem guarantees that this $\mathbf{X}_{proxy}$ minimizes the approximation error. The approximation error is quantified by the sum of the squares of the discarded singular values:
\vspace{-3mm}
\begin{align}
\|\mathbf{X}_{fg} - \mathbf{X}_{proxy}\|_F^2 = \sum_{i=K+1}^{N} \sigma_i^2.
\end{align}
Since $X_{fg}$ has a low effective rank, its singular values $\sigma_i$ decay rapidly. Therefore, if we select the fixed compression length $\mathbf{K}$ such that $\mathbf{K} \ll \mathbf{N}$, then the approximation error becomes negligible. This mathematically demonstrates that $\mathbf{X}_{proxy}$ can capture nearly all the essential information and underlying structure of the original high-dimensional, sparse $X_{fg}$, making the compression process \textit{information-nearly-lossless}. This provides a rigorous theoretical foundation for our learnable compression module.

\paragraph{SVD-based Dynamic Token Compression (DTC).} While SVD provides an optimal theoretical solution for compression, its non-differentiable nature prevents its direct use in an end-to-end framework. To bridge this gap, we propose SVD-Guided dynamic token compression module, which integrates the principled nature of SVD with the power of a learnable attention mechanism. First, we create a holistic token representation by infusing positional encoding, $\mathbf{X}'_{fg} = \mathbf{X}_{fg} + \mathbf{PE}_{orig}$, to ensure subsequent decisions are aware of both semantics and geometry while containing the spatial position information. Our dynamic compression iis then realized through the following process:
\begin{enumerate}
    \item \textbf{Online Rank Estimation:} For each input sequence $\mathbf{X}'_{fg} \in \mathbb{R}^{N \times C}$, we perform a fast, non-backpropagated SVD to obtain its singular values $\sigma_i$. Guided by the information bottleneck principle, we compute the cumulative energy of singular values and determine the $\textit{effective rank}$ $\mathbf{K}$ as the minimum number of components required to preserve a predefined threshold $\tau$ of the total variance. Specifically, $\mathbf{K}$ is derived by identifying the smallest integer such that: 
    \vspace{-1mm}
    \begin{align}
    \sum_{i=1}^\mathbf{K} \sigma_i^2 \geq \tau \sum_{j=1}^{\mathbf{N}} \sigma_j^2,
    \end{align}
    where $\tau$ is the energy retention threshold (see Appendix). This process dynamically yields two key outputs for each sample: the effective rank $\mathbf{K}$ and the corresponding optimal basis $\mathbf{Q}_{\text{SVD}}$ (the first $\mathbf{K}$ rows of $\mathbf{V}^T$). The rapid energy decay of its singular components concentrates the foreground's intrinsic information into a small set, enabling a high compression ratio while preserving nearly all essential information. Notably, this SVD step operates on the compact foreground feature and only computes the singular values, making it highly efficient. In practice, its latency is negligible (e.g., $<$1 ms on a 3090 GPU).

    \item \textbf{Dynamic Query Learning based Singular Value:} Our module maintains a fixed pool of $\mathbf{L}$ learnable \textit{Compression Queries} ($\mathbf{Q}_{\text{learn}} \in \mathbb{R}^{\mathbf{L} \times \mathbf{C}}$). We dynamically select the first $\mathbf{K}$ of these learnable queries from this pool.  This selection of the top-K queries encourages the model to learn on an ordered basis within the learnable query pool, where the first queries are trained to capture the most common and significant aspects of target objects. The final active queries, $\mathbf{Q}_{\text{act}} \in \mathbb{R}^{\mathbf{K} \times \mathbf{C}}$, are then formed by informing this learnable basis with the SVD prior:
    \begin{align}
        \mathbf{Q}_{\text{act}} = \mathbf{S}_K \mathbf{Q}_{\text{learn}} + \mathbf{Q}_{\text{SVD}}.
        \label{query_plus}
    \end{align}
    where $\mathbf{S}_K \mathbf{Q}_{\text{learn}}$ select the first $\mathbf{K}$ rows from the learnable query pool. This formulation allows the learnable queries $\mathbf{Q}_{\text{learn}}$ to act as task-specific adaptations on top of the strong, data-dependent prior provided by $\mathbf{Q}_{\text{SVD}}$.

    \item \textbf{Guided Cross-Attention:} The final compressed proxy token sequence, $\mathbf{X}_{p} \in \mathbb{R}^{\mathbf{K} \times \mathbf{C}}$, is produced by performing cross-attention between $\mathbf{K}$ active queries $\mathbf{Q}_{\text{act}} $ and the input tokens from $\mathbf{X}'_{fg}$:
    \begin{align}
    \mathbf{X}_{p} &= \mathbf{SoftM}\left(\frac{{\mathbf{Q}_{\text{act}}}\mathbf{W}_q (\mathbf{X}'_{fg}\mathbf{W}_k)^T}{\sqrt{C}}\right) \mathbf{X}'_{fg}\mathbf{W}_v.
\end{align}
\end{enumerate}

This hybrid approach is end-to-end trainable because the non-differentiable SVD step is only used to determine an integer index for slicing, while the gradient flows through the learnable queries and the cross-attention module. By doing so, our model's compression ratio is not determined by a learned heuristic but is dynamically guided by the intrinsic rank of the foreground representation. The effectiveness of this SVD-guided formulation is validated in our ablation studies (Tab.~\ref{ablation_query}), showing that combining the SVD prior with learnable queries outperforms using either component alone.

\subsubsection{Training with Adaptive Masking.}
In SVD-guided dynamic token compression module, a key challenge is that the number of active tokens, $\mathbf{K}$, is dynamically determined for each sample, while efficient batch training requires fixed-size tensors. To resolve this, we adopt am \textit{Adaptive Masking} strategy where tensor dimensions remain fixed at the maximum length, $\mathbf{L}$, during training, but only the dynamically selected tokens contribute to the final loss. The process is as follows:
\textbf{(1)} \textit{Adaptive $\mathbf{K}$ Determination:} In each forward pass, the effective rank $\mathbf{K}$ for each sample in the batch is determined via our online SVD estimation. \textbf{(2)} \textit{Mask Generation} A binary mask $\mathbf{M} \in \{0, 1\}^{\mathbf{L}}$ is generated for each sample, where the first $\mathbf{K}$ entries are 1 and the rest are 0. \textbf{(3)} \textit{Mask Application} This mask is passed as the subsequent \texttt{token\_mask} to all subsequent self-attention layers. This logically nullifies the contribution of the ''inactive'' queries (from $\mathbf{K+1}$ to $\mathbf{L}$) by ensuring their attention weights become zero after the softmax operation. This approach allows the gradient to flow only through the $\mathbf{K}$ active queries that were adaptively selected for that specific sample, while maintaining a consistent tensor shape for batch processing.

\paragraph{Prediction Head and Training Loss.} Adopting parallel regression branches from~\cite{synctrack,zhao2024ost}, we directly predict target parameters $(x,y,z,\theta)$ using the final backbone features, eliminating multi-scale aggregation used in~\cite{osp2b,corpnet}. The entire CompTrack network is trained end-to-end with a composite loss function that supervises both the foreground prediction and the final tracking task: The total loss, $\mathbf{L}_{total}$, is a weighted sum of two main components: \textbf{(1) Prediction Loss} ($\mathbf{L}_{pred}$): As described in Eq~\ref{equ:predictor_train}, this loss supervises the Spatial Foreground Predictor (SFP). \textbf{(2) Tracking Loss ($\mathbf{L}_{track}$)}: This loss supervises the final prediction head that operates on the compressed proxy tokens. It consists of a classification component to identify the target and a regression component to refine the 3D bounding box. The training loss is adopted from~\cite{v2b}:
$\mathbf{L}_{track}=\lambda_1\mathcal{L}_{(x,y)}+\lambda_2\mathcal{L}_{z}+\lambda_3\mathcal{L}_{rot}$, where $\lambda_1$, $\lambda_2$ and $\lambda_3$ are hyper-parameters to balance different losses. Details can refer to~\cite{v2b}. The final loss is formulated as:
\begin{align}
\mathbf{L}_{total} = \theta_1 \mathbf{L}_{pred} + \theta_2 \mathbf{L}_{track},
\end{align}
where 
$\theta_1$ and $\theta_2$ are hyper-parameters that balance the two tasks. Note that our SVD-guided compression module does not require a separate sparsity regularizer, as the compression ratio is directly guided by the data's intrinsic rank.

%% file: sec/4_experiments.tex
\begin{table*}[!ht]
\centering
    \resizebox{0.95\textwidth}{!}{
    \normalsize
    \begin{tabular}{cc|ccccc|ccc}
          \toprule[0.4mm]
            & & Car & Pedestrian &  Van & Cyclist& Mean & \multicolumn{3}{c}{Computation Efficiency} \\
           \multirow{-2}{*}{Tracker} & \multirow{-2}{*}{Publish} & [6,424]&[6,088] & [1,248] & [308] & [14,068]& FLOPs & FPS & Device \\
          \midrule
          SC3D~\cite{sc3d}& CVPR'19 &41.3 / 57.9   & 18.2 / 37.8 & 40.4 / 47.0 & 41.5 / 70.4 & 31.2 / 48.5 & 19.80 G &2&GTX 1080Ti \\
          P2B~\cite{p2b}& CVPR'20& 56.2 / 72.8 & 28.7 / 49.6 & 40.8 / 48.4 & 32.1 / 44.7 & 42.4 / 60.0 & 4.30 G &40&GTX 1080Ti \\
        PTT~\cite{ptt}&IROS'21 &67.8 / 81.8& 44.9 / 72.0 &43.6 / 52.5& 37.2 / 47.3 & 55.1 / 74.2& -& 40&GTX 1080Ti \\
        LTTR~\cite{lttr}&BMVC'21& 65.0 / 77.1 &33.2 / 56.8 & 35.8 / 45.6 &66.2 / 89.9& 48.7 / 65.8 & -& 23 &GTX 1080Ti  \\
        MLVSNet~\cite{mlvsnet}&ICCV'21 &56.0 / 74.0   & 34.1 / 61.1 & 52.0 / 61.4 & 34.4 / 44.5 & 45.7 / 66.6 &  -& 70 &GTX 1080Ti \\
         BAT~\cite{bat}&ICCV'21& 60.5 / 77.7 &42.1 / 70.1 & 52.4 / 67.0 &33.7 / 45.4& 51.2 / 72.8 & 2.77 G& 57 &RTX 2080  \\
        V2B~\cite{v2b}&NeurIPS'21 &70.5 / 81.3   & 48.3 / 73.5 & 50.1 / 58.0 & 40.8 / 49.7 & 58.4 / 75.2 & 5.57 G &37&TITAN RTX\\
        PTTR~\cite{pttr}& CVPR'22&65.2 / 77.4   & 50.9 / 81.6 & 52.5 / 61.8 & 65.1 / 90.5 & 57.9 / 78.2 & 2.61 G&50 &Tesla V100 \\
        M$^2$Track~\cite{m2track} & CVPR'22&65.5 / 80.8   & 61.5 / 88.2 & 53.8 / 70.7 & 73.2 / 93.5 & 62.9 / 83.4 & 2.54 G&57 &Tesla V100\\
        STNet~\cite{stnet} & ECCV'22 &72.1 / 84.0 &49.9 / 77.2& 58.0 / 70.6& 73.5 / 93.7& 61.3 / 80.1& 3.14 G& 35 &TITAN RTX\\
        CMT~\cite{cmt} & ECCV'22 &70.5 / 81.9 &49.1 / 75.5& 54.1 / 64.1& 55.1 / 82.4 &59.4 / 77.6 & - & 32&GTX 1080Ti\\
        GLT-T~\cite{glt}&AAAI'23 &68.2 / 82.1   & 52.4 / 78.8 & 52.6 / 62.9 & 68.9 / 92.1 & 60.1 / 79.3 & 3.87 G&30 &GTX 1080Ti \\
        OSP2B~\cite{osp2b}& IJCAI'23&67.5 / 82.3   & 53.6 / 85.1 & 56.3 / 66.2 & 65.6 / 90.5 & 60.5 / 82.3 & 2.57 G & 34& GTX 1080Ti \\
        CXTrack~\cite{cxtrack} & CVPR'23& 69.1 / 81.6 &67.0 / 91.5& 60.0 / 71.8& 74.2 / 94.3& 67.5 / 85.3& 4.63 G& 34&RTX 3090 \\
        MBPTrack~\cite{mbptrack} & ICCV'23 & 73.4 / 84.8 &68.6 / 93.9 &61.3 / 72.7& \underline{76.7} / 94.3& 70.3 / 87.9 & 2.88 G & 50&RTX 3090  \\
        SyncTrack~\cite{synctrack} &ICCV'23 &73.3 / 85.0 & 54.7 / 80.5&60.3 / 70.0& 73.1 / 93.8 & 64.1 / 81.9& 2.51 G & 45&TITAN RTX\\
          M$^2$Track++~\cite{m2track++} & TPAMI'23&71.1 / 82.7  & 61.8 / 88.7 & 62.8 / 78.5 & 75.9 / 94.0 & 66.5 / 85.2 & 2.54 G &57& Tesla V100\\
        SCVTrack~\cite{robust}&AAAI'24&68.7 / 81.9 &62.0 / 89.1& 58.6 / 72.8& \textbf{77.4} / 94.4& 65.2 / 84.6&-& 31&RTX 3090\\
        PTTR++~\cite{pttr++}& TPAMI'24&73.4 / 84.5 &55.2 / 84.7 &55.1 / 62.2& 71.6 / 92.8& 63.9 / 82.8& - & 43 &Tesla V100 \\
          OST~\cite{zhao2024ost} &TMM'24 &72.0 / 84.2 & 51.4 / 82.6&57.5 / 68.2& 49.2 / 60.4 & 61.3 / 81.6& 2.35 G & 47&TITAN RTX\\
         P2P$^\dagger$~\cite{p2p} & IJCV'25&\textbf{73.6} / \textbf{85.7} & \textbf{69.6} / \underline{94.0}&  \textbf{70.3} /  \textbf{83.9}& 75.5 / \underline{94.6} & \textbf{71.7} / \textbf{89.4}& 1.23 G& 65 & RTX 3090\\
         \midrule
         \rowcolor{myblue!18} \textbf{CompTrack} \textbf{(Ours)} & - & \underline{73.4} / \underline{85.2}&\underline{69.5} / \textbf{94.7} &\underline{68.5} / \underline{82.5}&76.0 / \textbf{94.8}&\underline{71.4} / \underline{89.3} & \textbf{0.94 G}&\textbf{90} & RTX 3090\\
          \bottomrule[0.4mm]
    \end{tabular}}
\vspace{-1mm}
\caption{Comparisons with state-of-the-art methods on KITTI dataset~\cite{kitti}. The upper and lower parts include two-stream and one-stream trackers, respectively. \textit{Success} / \textit{Precision} are used for evaluation. \textbf{Bold} and \underline{underline} denote the best result and the second-best one, respectively. $\dagger$ means our reimplementation based on official code.}
\label{table1}
\end{table*}

\begin{table*}[!htbp]
\centering
    \resizebox{0.95\textwidth}{!}{
    \normalsize
    \begin{tabular}{c|ccccc|c}
          \toprule[0.4mm]
           Tracker & Car [64,159] & Pedestrian [33,227]&  Truck [13,587] & Trailer [3,352] & Bus [2,953] & Mean [117,278] \\
          \midrule
          SC3D~\cite{sc3d} &22.31  / 21.93& 11.29  / 12.65& 30.67  / 27.73& 35.28  / 28.12& 29.35  / 24.08& 20.70  / 20.20\\
           P2B~\cite{p2b} & 38.81  / 43.18 &28.39  / 52.24& 42.95  / 41.59& 48.96  / 40.05& 32.95  / 27.41& 36.48  / 45.08\\
           PTT~\cite{ptt}  &41.22  / 45.26& 19.33  / 32.03&50.23  / 48.56&51.70  / 46.50&39.40  / 36.70 & 36.33  / 41.72 \\
           BAT~\cite{bat} & 40.73  / 43.29 &28.83  / 53.32& 45.34  / 42.58& 52.59  / 44.89 &35.44  / 28.01 &38.10  / 45.71\\
           M$^2$Track~\cite{m2track} &55.85  / 65.09& 32.10  / 60.92 &57.36  / 59.54& 57.61  / 58.26& 51.39  / 51.44& 49.23  / 62.73\\ 
           PTTR~\cite{pttr}& 51.89  / 58.61& 29.90  / 45.09& 45.30  / 44.74 &45.87  / 38.36& 43.14  / 37.74& 44.50  / 52.07\\
           GLT-T~\cite{glt} &48.52  / 54.29&31.74  / 56.49& 52.74  / 51.43&57.60  / 52.01& 44.55  / 40.69&44.42  / 54.33 \\
           PTTR++~\cite{pttr++} & 59.96 / 66.73& 32.49 / 50.50 &59.85 / 61.20 &54.51 / 50.28 &53.98 / 51.22&51.86 / 60.63\\
           MBPTrack~\cite{mbptrack}  & 62.47  /  70.41 & 45.32  /  74.03 &  62.18  /  63.31 &  65.14  /  61.33&   55.41  /  51.76 & 57.48  /  69.88\\
           P2P$^\dagger$~\cite{p2p}& \underline{64.61} / \underline{71.98} & \underline{45.64} / \underline{74.62} & \underline{64.42} / \underline{65.37} & \underline{70.23} / \underline{66.08} & \underline{58.54} / \underline{56.13} &\underline{59.22} / \underline{71.19} \\
            \midrule
            \rowcolor{myblue!18} \textbf{CompTrack (Ours)} &   \textbf{65.70}  /  \textbf{73.50}&  \textbf{47.86}  /  \textbf{77.52}& \textbf{68.19}  /  \textbf{69.78}& \textbf{72.89}  /  \textbf{68.11}& \textbf{61.74}  /  \textbf{58.88}&  \textbf{61.04}  /  \textbf{73.68}\\
          \bottomrule[0.4mm]
  \end{tabular}}
\vspace{-1mm}
\caption{Comparisons with state-of-the-art methods on nuScenes dataset~\cite{nuScenes}. \textit{Success} / \textit{Precision} are used for evaluation. \textbf{Bold} and \underline{underline} denote the best result and the second-best one, respectively.}
\label{table3}
\vspace{-4mm}
\end{table*}
\vspace{-0.1mm}
\section{Experiments}

\noindent\textbf{Implementation Details.}
We follow the common setup~\cite{p2b,m2track} and conduct extensive experiments on KITTI~\cite{kitti}, nuScenes~\cite{nuScenes} and Waymo Open Dataset (WOD)~\cite{waymo}. The evaluation metrics is followed the common setup~\cite{ptt,glt,mbptrack} to report \textit{Success} and \textit{Precision} based on one pass evaluation (OPE)~\cite{otb2013,kristan2016novel}. More implementation details are in appendix.

\vspace{-2mm}
\subsection{Comparison with State-of-the-art Trackers}
\noindent\textbf{Results on KITTI.} We present comprehensive accuracy and speed comparisons on KITTI~\cite{kitti} dataset. As shown in Tab.~\ref{table1}, CompTrack achieves a highly competitive mean Success/Precision of 71.4\% / 89.3\%, ranking it among the top performers. Its key advantage lies in exceptional efficiency: while matching the accuracy of leading methods like P2P, our model operates at a high speed of 90 FPS with significantly lower FLOPs. This speedup is a direct result of our IB-DTC design, which avoids the costly processing of redundant data by distilling the foreground into a compact set of low-rank tokens, proving the design to be both powerful and efficient.

\noindent\textbf{Results on nuScenes.} On the challenging, sparser and large-scale nuScenes~\cite{nuScenes} dataset, Tab.~\ref{table3} shows that our CompTrack sets a new state-of-the-art with a mean Success/Precision of 61.04\% / 73.68\%, outperforming all prior trackers across all categories. This strong performance in nuScenes' sparse and complex environments underscores the robustness of our core design. By dynamically filtering and compressing the foreground into a compact, low-rank token representation, CompTrack maintains high accuracy where other methods falter, validating CompTrack's potential for large-scale practical deployment.

\begin{table*}[!htbp]
\centering
    \resizebox{0.95\textwidth}{!}{
    \begin{tabular}{c|c|cccc|cccc}
          \toprule[0.4mm]
         & &\multicolumn{4}{c}{Vehicle} & \multicolumn{4}{c}{Pedestrian}\\
         & & Easy& Medium & Hard & Mean & Easy & Medium & Hard & Mean \\
        \multirow{-3}{*}{Tracker} & \multirow{-3}{*}{Mean} & [67,832]&[61,252]&[56,647]&[185,731]& [85,280]&[82,253]&[74,219]&[241,752] \\
          \midrule[0.4mm]
          P2B~\cite{p2b}& 33.0 / 43.8 & 57.1 / 65.4 & 52.0 / 60.7 & 47.9 / 58.5 & 52.6 / 61.7 & 18.1 / 30.8 & 17.8 / 30.0 & 17.7 / 29.3 & 17.9 / 30.1 \\ 
          BAT~\cite{bat}& 34.1 / 44.4 & 61.0 / 68.3 & 53.3 / 60.9 & 48.9 / 57.8 & 54.7 / 62.7 & 19.3 / 32.6 & 17.8 / 29.8 & 17.2 / 28.3 & 18.2 / 30.3 \\
          V2B~\cite{v2b}& 38.4 / 50.1 & 64.5 / 71.5 & 55.1 / 63.2 & 52.0 / 62.0 & 57.6 / 65.9 & 27.9 / 43.9 & 22.5 / 36.2 & 20.1 / 33.1 & 23.7 / 37.9 \\
         STNet~\cite{stnet}&40.4 / 52.1 & {65.9} / {72.7}& {57.5} / {66.0}& {54.6} / {64.7} &{59.7} / {68.0}& 29.2 / 45.3& 24.7 / 38.2& 22.2 / 35.8 &25.5 / 39.9  \\
         M$^2$Track~\cite{m2track}&44.6 / 58.2&68.1 / 75.3& 58.6 / 66.6 &55.4 / 64.9 &61.1 / 69.3& 35.5 / 54.2 &30.7 / 48.4 &29.3 / 45.9& 32.0 / 49.7 \\
    CXTrack~\cite{cxtrack} & 42.2 / 56.7 & 63.9 / 71.1 &54.2 / 62.7& 52.1 / 63.7 &57.1 / 66.1& 35.4 / 55.3 &29.7 / 47.9& 26.3 / 44.4 &30.7 / 49.4 \\
    MBPTrack~\cite{mbptrack} &46.0 / 61.0 & \textbf{68.5} / \textbf{77.1} &\textbf{58.4} / \textbf{68.1}& \textbf{57.6} / \textbf{69.7}& \textbf{61.9} / \textbf{71.9} &37.5 / 57.0 &33.0 / 51.9& 30.0 / 48.8& 33.7 / 52.7 \\
    
    P2P$^\dagger$~\cite{p2p} &\underline{47.2} / \underline{62.9} &  66.2 / 73.8 & 57.8 / 67.0 &56.8 / 68.1 & 60.0 / 69.1 & \underline{43.7} / \underline{65.2} & \underline{36.4} / \underline{57.1} & \underline{31.3} / \underline{51.0} & \underline{37.4} / \underline{58.1}\\
    \midrule[0.4mm]
    \rowcolor{myblue!18} \textbf{CompTrack (Ours)}  &  \textbf{48.6} / \textbf{65.7}&\underline{68.2} / \underline{72.8}&\underline{57.4} / \underline{66.0} &\underline{57.2} / \underline{69.4} & \underline{61.2} / \underline{69.6} &\textbf{43.8} / \textbf{68.6}&\textbf{38.7} / \textbf{62.6} &\textbf{33.8} / \textbf{56.0} &\textbf{39.0} / \textbf{62.7}\\
    \bottomrule[0.4mm]
    \end{tabular}}
\vspace{-1mm}
\caption{Comparisons with state-of-the-art methods on Waymo Open Dataset~\cite{waymo}. \textit{Success} / \textit{Precision} are used for evaluation. \textbf{Bold} and \underline{underline} denote the best result and the second-best one, respectively.}
\label{table2}
\end{table*}

\begin{table*}[!htbp]
\centering
\resizebox{0.95\linewidth}{!}{
\small
\begin{tabular}{c|cc|c|c|ccccc}
  \toprule[0.4mm]
  Setting&SFP & IB-DTC & FPS & Mean & Car & Pedestrian & Truck & Trailer &Bus  \\
  \midrule
  (A) &\xx  & \xx &48 &59.38 / 71.63 & 64.73 / 71.90 & 45.70 / 74.98 & 64.86 / 65.80 & 70.67 / 66.56 & 59.30 / 56.60 \\
  (B) &\cc & \xx & 55 & 60.01 / 72.20 & 65.01 / 72.25 & 46.21 / 75.88 & 67.50 / 67.51 & 71.05 / 67.03 & 59.55 / 57.12 \\
  (C) &\xx & \cc & 75 & 59.95 / 72.18 & 65.01 / 72.25 & 46.21 / 75.88 & 65.50 / 66.51 & 71.05 / 67.03 & 59.55 / 57.12 \\
  (D) &\cc & \cc & 90 &$\textbf{61.04}$ / $\textbf{73.68}$ & $\textbf{65.70}$ / $\textbf{73.50}$ & $\textbf{47.86}$ / $\textbf{77.52}$ & $\textbf{68.19}$ / $\textbf{69.78}$ & $\textbf{72.89}$ / $\textbf{68.11}$ & $\textbf{61.74}$ / $\textbf{58.88}$\\
  \bottomrule
\end{tabular}
}
\caption{Ablation of proposed SFP and IB-DTC design. Baseline from the reproduced P2P~\cite{p2p} with replacing the voxel-based backbone with self-attention~\cite{swin}.}
\vspace{-1mm}
\label{table_ablation}
\vspace{-3mm}
\end{table*}

\noindent\textbf{Results on Waymo.} We follow common setup~\cite{m2track,v2b,p2p} and test the generalization of CompTrack by evaluating our KITTI-trained models directly on the large-scale WOD~\cite{waymo} dataset, with results in Tab.~\ref{table2}. CompTrack demonstrates superior generalization, particularly in the Pedestrian category, where it establishes a new state-of-the-art with a mean Success/Precision of 39.0\% / 62.7\%. While memory-based methods like MBPTrack show an edge on Vehicle tracking at the cost of speed, our method's strong performance on pedestrians highlights the robustness of our dynamic token compression. This mechanism learns to distill the most critical features from sparse inputs, ensuring excellent generalization to unseen data and difficult object classes.

\vspace{-1mm}
\subsection{Ablation Study}

Due to the limited scale and diversity of the KITTI, to comprehensively evaluate our method, we follow the recent protocol~\cite{hu2025mvctrack, p2p} and perform ablations in the large-scale nuScenes~\cite{nuScenes} dataset. More ablation studies can refer to Appendix.

\noindent\textbf{Ablation of SFP and IB-DTC Designs.} We conduct ablation studies to validate the performance of each proposed design. As shown in Tab.~\ref{table_ablation}, our baseline model (A), which disables both modules, achieves a mean Success/Precision of 59.38\%/71.63\% at 48 FPS. By individually integrating the Spatial Foreground Predictor (SFP) in setting (B) or the Information Bottleneck-guided Dynamic Token Compression (IB-DTC) in setting (C), both accuracy and speed are notably improved. This confirms the distinct benefits of filtering spatial redundancy and compressing informational redundancy. Our full model (D), which combines both modules, achieves the best performance and the highest speed (90 FPS), demonstrating that they are complementary.

\begin{table}[!htbp]
\centering
\resizebox{\linewidth}{!}{
\begin{tabular}{l|c|c|cc}
\toprule[0.4mm]
\textbf{Query Formulation} & \textbf{FPS} & \textbf{Mean} & \textbf{Car} & \textbf{Pedestrian} \\
\midrule
(A) Learnable-Only & 88 & 60.70 / 73.25 & 65.31 / 73.02 & 47.45 / 76.91 \\
(B) SVD-Only & \textbf{91} & 60.15 / 72.50 & 64.95 / 72.33 & 46.91 / 75.85 \\
\midrule[0.4mm]
(C) IB-DTC (Ours) & 90 & \textbf{61.04} / \textbf{73.68} & \textbf{65.70} / \textbf{73.50} & \textbf{47.86} / \textbf{77.52} \\
\bottomrule[0.4mm]
\end{tabular}
}
\vspace{-1mm}
\caption{Ablation study of our SVD-guided query formulation. We compare our full hybrid model against variants that use only learnable queries or only the SVD-derived basis.}
\vspace{-3mm}
\label{ablation_query}
\end{table}

\begin{figure}[!htbp]
\centering
\includegraphics[width=0.99\linewidth]{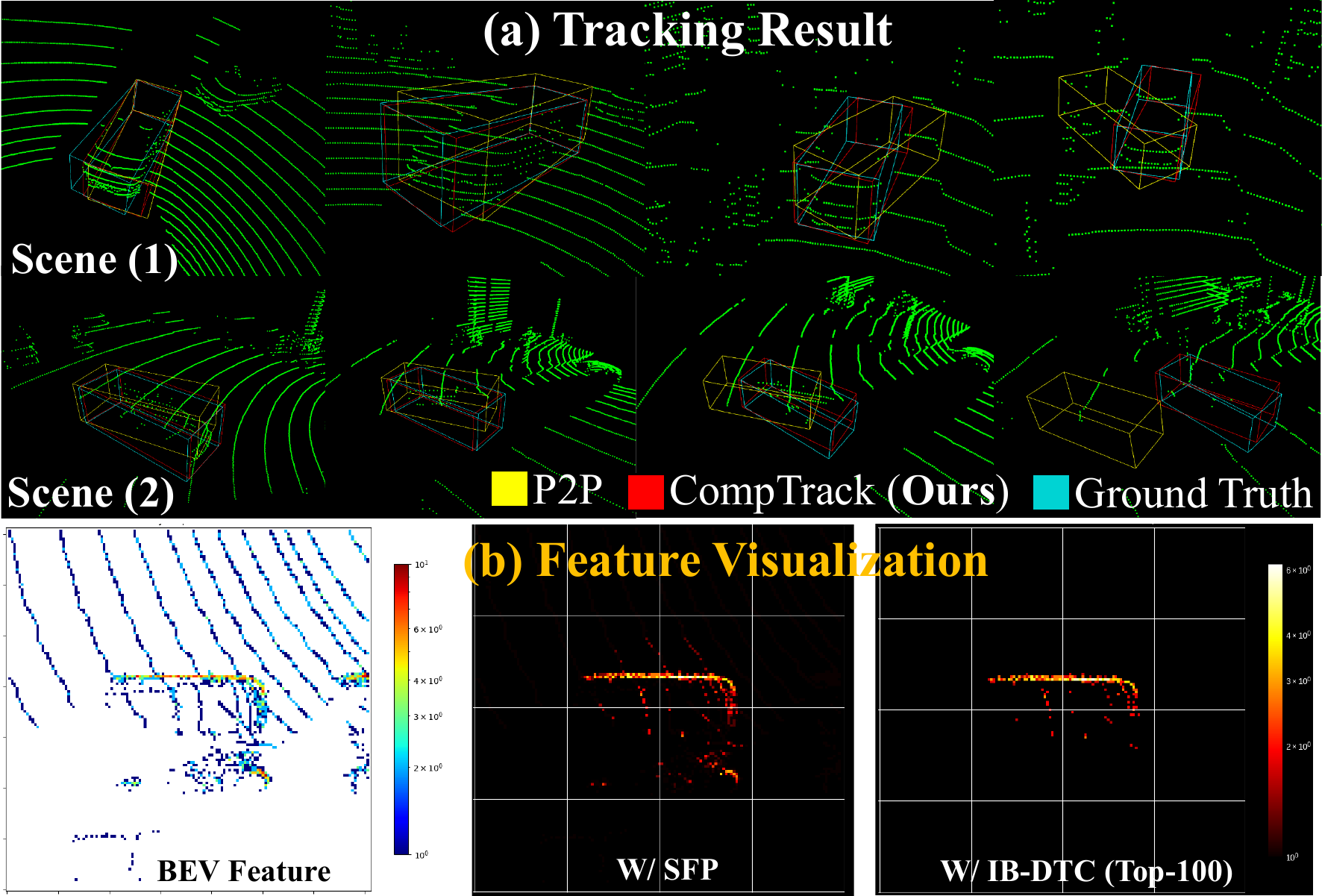}
\vspace{-1mm}
\caption{The visualization of tracking and feature maps.}
\label{vis_fig}
\vspace{-7mm}
\end{figure}

\noindent\textbf{Ablation of SVD-guided Token Generation.} 
We ablate our dynamic token compression in Table~\ref{ablation_query}. The \textit{Learnable-Only} (A) establishes a strong baseline, demonstrating the capability of end-to-end trained queries. Using the non-learnable \textit{SVD-Only} (B) is also effective, confirming the benefits of SVD priors. However, our (C) \textit{IB-DTC}, which synergizes both, achieves the best performance. This confirms that the SVD provides a meaningful contribution by guiding the learnable queries with a strong prior.

\noindent\textbf{Feature and Tracking Visualization.} 
As shown in Fig.~\ref{vis_fig}, we visualize tracking results on the nuScenes~\cite{nuScenes}. It can be seen that whether in dense (Fig.~\ref{vis_fig} (a-1)), or sparse scenarios (Fig.~\ref{vis_fig} (a-2)), our CompTrack is able to tightly track the target compared to previous SOTA P2P~\cite{p2p}, which highlights the CompTrack's superior performance. Additionally, we plot the feature maps with SPF and IB-DTC designs ((Fig.~\ref{vis_fig} (b))). The visualization results show the effectiveness of proposed CompTrack.

%% file: sec/5_conclusion.tex
\vspace{-2mm}
\section{Conclusion}
In this paper, we propose \textbf{CompTrack}, a novel end-to-end framework for 3D SOT that tackles the challenges of spatial and informational redundancy inherent to point clouds. Our CompTrack first employs a Spatial Foreground Predictor to filter background noise, followed by an Information Bottleneck-guided Dynamic Token Compression module that adaptively compresses the foreground into a low-rank representation. By reducing both forms of redundancy, our method achieves an exceptional balance between accuracy and efficiency. Extensive experiments demonstrate that CompTrack has leading performance on three well-known benchmarks, striking a better trade-off between efficiency and accuracy in latency-sensitive tracking tasks. 

\vspace{-2mm}
\section{Future Work}
We believe that the proposed dual-redundancy paradigm in CompTrack, tackling both spatial and informational redundancy, offers a powerful principle for designing high-efficiency 3D perception models. Our Information Bottleneck-guided adaptive compression, in particular, presents an effective strategy for handling complex point cloud representations in latency-sensitive tracking tasks. However, CompTrack still struggles in extremely sparse scenes, where the object is partially visible. For future work, this framework could be extended by incorporating multi-frame temporal information~\cite{fan2025beyond} to better model object motion or by fusing it with dense RGB data~\cite{hu2025mvctrack} to leverage complementary appearance cues for even greater robustness and accuracy. Besides, we believe that model quantization techniques~\cite{zhou2024lidarptq,gsq,wang2025point4bit,xumambaquant,hu2025ostquant,yu2025mquant} are orthogonal to further improve model efficiency and investigate the integration with privacy-preserving federated learning~\cite{liu2024fedbcgd,liu2025consistency,liu2025improving}.

\vspace{-1mm}
\section{Acknowledgement}
This work was supported by the National Natural Science Foundation of China (No.62271143), Frontier Technologies R\&D Program of Jiangsu (No. BF2024060) and the Big Data Computing Center of Southeast University.

%% file: sec/6_supp.tex
\appendix
\section{Appendix}

\subsection{More Experiments}

\begin{table}[!htbp]
\centering
\label{table:ablation_tau}
\resizebox{1.0\linewidth}{!}{
\begin{tabular}{c|c|cc}
\toprule[0.4mm]
\textbf{Threshold ($\tau$)} & \textbf{Avg. $K_{\text{eff}}$} & \textbf{Success (\%)} & \textbf{Precision (\%)} \\
\midrule
0.95  & $\approx 65$  & 59.88 & 72.16 \\
\textbf{0.99} & $\approx \textbf{78}$ & \textbf{61.04} & \textbf{73.68} \\
0.999 & $\approx 125$ & 61.05 & 73.68 \\
\bottomrule[0.4mm]
\end{tabular}
}
\caption{Sensitivity analysis of the energy retention threshold $\tau$ on the nuScenes dataset. Our chosen value is highlighted in bold.}
\label{table:ablation_tau}
\vspace{-2mm}
\end{table}

\subsubsection{Ablation of Energy Retention Threshold in SVD.} 

Here, we study the effect of the energy retention threshold $\tau$ in Table~\ref{table:ablation_tau}. While a loose threshold ($\tau=0.95$) degrades performance due to significant information loss, the model is not sensitive to the specific choice within the high-fidelity range of [0.99-0.999]. This demonstrates the robustness of our approach and validates our default setting of $\tau=0.99$.

\begin{table}[!htbp]
\centering

\label{table:ablation_fusion}
\resizebox{1.0\linewidth}{!}{
\begin{tabular}{l|c|cc}
\toprule[0.4mm]
\textbf{Fusion Method} & \textbf{Formula} & \textbf{Success (\%)} & \textbf{Precision (\%)} \\
\midrule
Learnable-Only & $\mathbf{Q}_{\text{act}} = \mathbf{Q}_{\text{learn}}$ & 60.70 & 73.25 \\
SVD-Only & $\mathbf{Q}_{\text{act}} = \mathbf{Q}_{\text{learn}}$ & 60.15 & 72.50 \\
\midrule[0.2mm]
Concat+ Linear & $\mathbf{Q}_{\text{act}} = \text{Linear}(\text{cat}[\mathbf{Q}_{\text{learn}}, \mathbf{Q}_{\text{SVD}}])$ & 60.85 & 73.40 \\
\textbf{Addition (Ours)} & $\mathbf{Q}_{\text{act}} = \mathbf{Q}_{\text{learn}} + \mathbf{Q}_{\text{SVD}}$ & \textbf{61.04} & \textbf{73.68} \\
\bottomrule[0.4mm]
\end{tabular}
}
\caption{Ablation study on the query fusion strategy. Our proposed additive fusion demonstrates the most effective performance.}
\label{table:ablation_fusion}
\vspace{-2mm}
\end{table}

\subsubsection{Ablation Study on Query Fusion Strategy}
\label{sec:appendix_fusion}

In our IB-DTC module, we propose combining the learnable queries ($\mathbf{Q}_{learn}$) with the SVD prior ($\mathbf{Q}_{SVD}$) to form the final active queries ($\mathbf{Q}_{act}$). The method of fusing these two components is a critical design choice. To validate our additive formulation ($\mathbf{Q}_{\text{act}} = \mathbf{Q}_{\text{learn}} + \mathbf{Q}_{\text{SVD}}$), which treats the learnable component as a task-specific residual, we compare it against two strong alternatives:

\begin{itemize}
    \item \textbf{Learnable-Only:} This baseline uses only the learnable queries, forgoing the SVD prior entirely ($\mathbf{Q}_{act} = \mathbf{Q}_{learn}$).
    \item \textbf{SVD-Only:} This baseline uses only the SVD priors, ($\mathbf{Q}_{act} = \mathbf{Q}_{SVD}$).
    \item \textbf{Concatenation-based Fusion:} This method concatenates the two components along the feature dimension and uses a learnable linear layer to project them back to the original dimension, allowing the model to learn an optimal fusion strategy.
\end{itemize}

The results on the nuScenes dataset are presented in Table~\ref{table:ablation_fusion}. Both methods that incorporate the SVD prior outperform the `Learnable-Only` and `SVD-Only` baseline, confirming the value of the data-driven guidance. Notably, our proposed additive fusion achieves the best performance, slightly surpassing the more complex concatenation-based approach. This additive formulation is inspired by residual learning principles, allowing the learnable queries to act as task-specific adaptations on top of the strong, data-dependent prior provided by SVD. This validates our hypothesis that formulating the learnable component as a residual on top of the strong SVD prior is a more effective and direct way to integrate these two sources of information.

\begin{table}[h]
\centering
\resizebox{1.0\linewidth}{!}{
\begin{tabular}{lcc}
\toprule
Method & Success (\%) & FPS \\
\midrule
Uniform 1/8 grid  & 58.77 & 92 \\
Random Drop 75\%  & 58.45 & 93 \\
Fixed‑K TokenLearner ($K\!=\!128$) & 59.21 & 86 \\
\textbf{IB‑DTC (ours, $\bar{K}\!\approx\!78$)} & \textbf{61.04} & 90 \\
\bottomrule
\end{tabular}
}
\caption{Efficiency/accuracy of alternative compression schemes on
nuScenes.  IB‑DTC outperforms all simpler baselines.}
\label{compbaseline}
\vspace{-2mm}
\end{table}

\subsubsection{Additional Compression Baselines}
We further compare IB-DTC with three lightweight but naïve compression baselines. As Table~\ref{compbaseline} shows, neither geometry-agnostic subsampling (uniform / random) nor a strong 2D TokenLearner~\cite{ryoo2021tokenlearner} (fixed K) can match the speed-accuracy trade-off of our rank-aware, IB-guided dynamic token compressor.
This validates the necessity of coupling online rank estimation with learnable residual queries rather than relying on heuristic token reduction

\subsection{More Implementation Details}
\vspace{-1mm}
\subsubsection{Dataset and Evaluation Metrics} 

Our method's performance is validated on three large-scale benchmarks: KITTI~\cite{kitti}, nuScenes~\cite{nuScenes} and Waymo Open Dataset (WOD)~\cite{waymo}. The evaluation protocol adheres to established conventions~\cite{p2b,m2track,mbptrack}: models are trained and tested on both the KITTI and nuScenes datasets, while the model pre-trained on KITTI is used for generalization testing on WOD. Among these, nuScenes and WOD represent more challenging scenarios due to their substantial data volume and complex environments. Furthermore, the data collection hardware differs; KITTI and WOD utilize 64-beam LiDAR sensors, whereas nuScenes employs a 32-beam LiDAR, resulting in comparatively sparser point cloud data.

\noindent\textbf{KITTI.} The KITTI benchmark~\cite{kitti} provides 21 sequences for training and 29 for testing. As the ground truth for the official test set is withheld for evaluation server purposes, we adopt the widely-used data partition~\cite{sc3d,p2b,bat,m2track,cxtrack,mbptrack,cutrack} that divides the training sequences into a new training set (sequences 0-17), validation set (18-19), and test set (20-21).

\noindent\textbf{NuScenes.} The nuScenes dataset~\cite{nuScenes} is composed of 1000 scenes, officially divided into 700 for training, 150 for validation, and 150 for testing. In accordance with prior works~\cite{bat,m2track,mbptrack}, our CompTrack model is trained on the ``train$\_$track'' subset, and its performance is reported on the validation set.

\noindent\textbf{Waymo.} The Waymo Open Dataset (WOD) contains 1121 distinct tracklets for evaluation. These tracklets are stratified into three difficulty levels (easy, medium, and hard) based on the point cloud density of the target object. Our evaluation on this benchmark adheres to the protocol established by LiDAR-SOT~\cite{lidarsot,v2b,stnet}.

\noindent\textbf{Evaluation Metrics.} The tracking performance is assessed using the standard One Pass Evaluation (OPE) protocol~\cite{otb2013,kristan2016novel}. The evaluation relies on two primary metrics: Success and Precision. The Success metric is determined by the Intersection over Union (IoU) between the predicted and ground-truth bounding boxes, while the Precision metric measures the Euclidean distance between the centers of the two boxes.

\subsubsection{Model Hyper-parameters}
The extended ranges of input point cloud regions are defined as extended range [(-4.8,4.8),(-4.8,4.8),(-1.5,1.5)] and [(-1.92,1.92),(-1.92,1.92),(-1.5,1.5)] to contain relevant points for cars and humans, respectively. The spatial resolution after pillar feature encoder is $128\times128\times1$. We first crop out point cloud regions from consecutive point clouds $\mathcal{P}_{t-1}$ and $\mathcal{P}_t$ that have the same aspect ratio as the target and are 2 times the target box size. Then, a resolution-consistent pillarization operation with spatial resolution of $128\times 128$ is applied to structurally normalize the cropped point clouds.

\subsubsection{Data Augmentation.}
During the training stage, we introduce simulated test errors for consecutive frames to enhance model's accuracy. In contrast to previous methods~\cite{p2b,m2track,mbptrack,cutrack} that crop point cloud regions in $t-1$ and $t$ frames centered on target boxes in the corresponding frame, respectively, we use the target box in $t-1$ frame as the center of the cropped point cloud regions in both $t-1$ and $t$ frame. In this way, the model is able to learn tracking errors that occurred in the testing stage. In addition, we randomly flip target points and box horizontally and rotate it uniformly around its longitudinal axes in the range [-5$^\circ$,5$^\circ$].

\noindent\textbf{Training.} 
We train our tracking models using the AdamW optimizer on a NVIDIA RTX 3090 GPU, with a batch size of 128. Following ~\cite{p2p}, the initial learning rate is set to 1e-4 and is decayed by a factor of 5 every 20 epochs. 

\noindent\textbf{Comptrack Inference Pipeline.}
During the inference stage, CompTrack continuously regresses out the final bbox frame-by-frame. The regressed results are then applied to the target box $\mathcal{B}_{t-1}$ in the previous frame to locate the target box $\mathcal{B}_{t}$ in the current frame. Alg.~\ref{alg:inference} presents the whole inference process of CompTrack in a point cloud sequence.

\begin{algorithm}[H]
\caption{Inference Process of CompTrack in a point cloud sequence}
\label{alg:inference}
\textbf{Input:} Initial template points $\mathcal{P}_1$ with box $\mathcal{B}_1$; Search area points for subsequent frames $\{\mathcal{P}_t\}_{t=2}^{T}$. \\
\textbf{Output:} Tracking results $\{\mathcal{B}_t\}_{t=2}^{T}$.
\begin{algorithmic}[1]
\STATE Extract template BEV features $\mathbf{F}_{\text{template}}$ from $\mathcal{P}_1$. \textcolor{gray}{\textit{// Initial Setup}}
\FOR{$t$ in 2 to $T$}
    \STATE Extract search area BEV features $\mathbf{F}_{s}$ from $\mathcal{P}_t$.
    
    \STATE \textcolor{blue}{\textit{// Stage 1: Filter Spatial Redundancy}}
    \STATE Predict foreground heatmap $\mathbf{Y}_{\text{pred}} = \text{SFP}(\mathbf{F}_{\text{template}}, \mathbf{F}_{s})$.
    \STATE Obtain refined foreground features $\hat{\mathbf{F}}_{s} = \mathbf{F}_s \odot \mathbf{Y}_{\text{pred}}$.
    
    \STATE \textcolor{blue}{\textit{// Stage 2: Compress Informational Redundancy}}
    \STATE Tokenize $\hat{\mathbf{F}}_{s}$ and add positional encoding to get $\mathbf{X}'_{fg}$.
    \STATE Generate proxy tokens $\mathbf{X}_{\text{proxy}} = \text{IB-DTC}(\mathbf{X}'_{fg})$.
    
    \STATE \textcolor{blue}{\textit{// Final Prediction}}
    \STATE Predict bounding box $\mathcal{B}_t = \text{PredictionHead}(\mathbf{X}_{\text{proxy}})$.
\ENDFOR
\STATE \textbf{Return} $\{\mathcal{B}_t\}_{t=2}^{T}$.
\end{algorithmic}
\end{algorithm}

%% file: aaai2026.bib
@article{javed2022visual,
  title={Visual object tracking with discriminative filters and siamese networks: a survey and outlook},
  author={Javed, Sajid and Danelljan, Martin and Khan, Fahad Shahbaz and Khan, Muhammad Haris and Felsberg, Michael and Matas, Jiri},
  journal={IEEE Transactions on Pattern Analysis and Machine Intelligence},
  year={2022},
  publisher={IEEE}
}

@inproceedings{hu2025mvctrack,
  title={Mvctrack: Boosting 3d point cloud tracking via multimodal-guided virtual cues},
  author={Hu, Zhaofeng and Zhou, Sifan and Yuan, Zhihang and Yang, Dawei and Zhao, Shibo and Liang, Ci-jyun},
  booktitle={2025 IEEE International Conference on Robotics and Automation (ICRA)},
  pages={3745--3751},
  year={2025},
  organization={IEEE}
}

@inproceedings{yu2025mquant,
  title={MQuant: Unleashing the Inference Potential of Multimodal Large Language Models via Static Quantization},
  author = {Yu, Jiangyong and Zhou, Sifan and Yang, Dawei and Li, Shuoyu and Wang, Shuo and Hu, Xing and Xu, Chen and Xu, Zukang and Shu, Changyong and Yuan, Zhihang},
  year={2025},
isbn = {9798400720352},
publisher = {Association for Computing Machinery},
address = {New York, NY, USA},
doi = {10.1145/3746027.3755433},
booktitle={Proceedings of the 33rd ACM International Conference on Multimedia},
location = {Dublin, Ireland},
}

@inproceedings{LoRAT,
  title={Tracking meets lora: Faster training, larger model, stronger performance},
  author={Lin, Liting and Fan, Heng and Zhang, Zhipeng and Wang, Yaowei and Xu, Yong and Ling, Haibin},
  booktitle={European Conference on Computer Vision},
  pages={300--318},
  year={2024},
  organization={Springer}
}

@book{golub2013matrix,
  title={Matrix computations},
  author={Golub, Gene H and Van Loan, Charles F},
  year={2013},
  publisher={JHU press}
}

@article{eckart1936approximation,
  title={The approximation of one matrix by another of lower rank},
  author={Eckart, Carl and Young, Gale},
  journal={Psychometrika},
  volume={1},
  number={3},
  pages={211--218},
  year={1936},
  publisher={Springer-Verlag}
}

@inproceedings{zhou2025pillarhist,
  title={Pillarhist: A quantization-aware pillar feature encoder based on height-aware histogram},
  author={Zhou, Sifan and Yuan, Zhihang and Yang, Dawei and Hu, Xing and Qian, Jian and Zhao, Ziyu},
  booktitle={Proceedings of the Computer Vision and Pattern Recognition Conference},
  pages={27336--27345},
  year={2025}
}

@inproceedings{mamba4d,
  author={Liu, Jiuming and Han, Jinru and Liu, Lihao and Aviles-Rivero, Angelica I. and Jiang, Chaokang and Liu, Zhe and Wang, Hesheng},
  booktitle={2025 IEEE/CVF Conference on Computer Vision and Pattern Recognition (CVPR)}, 
  title={Mamba4D: Efficient 4D Point Cloud Video Understanding with Disentangled Spatial-Temporal State Space Models}, 
  year={2025},
  pages={17626-17636},
  doi={10.1109/CVPR52734.2025.01642}
}

@inproceedings{dai2025unbiased,
  title={Unbiased Missing-modality Multimodal Learning},
  author={Dai, Ruiting and Li, Chenxi and Yan, Yandong and Mo, Lisi and Qin, Ke and He, Tao},
  booktitle={Proceedings of the IEEE/CVF International Conference on Computer Vision},
  pages={24507--24517},
  year={2025}
}

@inproceedings{yin2025knowledge,
  title={Knowledge-Aligned Counterfactual-Enhancement Diffusion Perception for Unsupervised Cross-Domain Visual Emotion Recognition},
  author={Yin, Wen and Wang, Yong and Duan, Guiduo and Zhang, Dongyang and Hu, Xin and Li, Yuan-Fang and He, Tao},
  booktitle={Proceedings of the Computer Vision and Pattern Recognition Conference},
  pages={3888--3898},
  year={2025}
}

@inproceedings{zhao2023ada3d,
  title={Ada3d: Exploiting the spatial redundancy with adaptive inference for efficient 3d object detection},
  author={Zhao, Tianchen and Ning, Xuefei and Hong, Ke and Qiu, Zhongyuan and Lu, Pu and Zhao, Yali and Zhang, Linfeng and Zhou, Lipu and Dai, Guohao and Yang, Huazhong and others},
  booktitle={Proceedings of the IEEE/CVF International Conference on Computer Vision},
  pages={17728--17738},
  year={2023}
}

@Article{jmse13091615,
AUTHOR = {Cao, Fuyu and Xu, Hongli and Ru, Jingyu and Li, Zhengqi and Zhang, Haopeng and Liu, Hao},
TITLE = {Collision Avoidance of Multi-UUV Systems Based on Deep Reinforcement Learning in Complex Marine Environments},
JOURNAL = {Journal of Marine Science and Engineering},
VOLUME = {13},
YEAR = {2025},
NUMBER = {9},
ARTICLE-NUMBER = {1615},
URL = {https://www.mdpi.com/2077-1312/13/9/1615},
ISSN = {2077-1312},
DOI = {10.3390/jmse13091615}
}

@article{ZHANG2025130845,
title = {Conditional variational underwater image enhancement with kernel decomposition and adaptive hybrid normalization},
journal = {Neurocomputing},
volume = {650},
pages = {130845},
year = {2025},
issn = {0925-2312},
doi = {https://doi.org/10.1016/j.neucom.2025.130845},
url = {https://www.sciencedirect.com/science/article/pii/S0925231225015176},
author = {Haopeng Zhang and Hongli Xu and Hao Liu and Xiaosheng Yu and Xiangyue Zhang and Chengdong Wu},
}

@inproceedings{ScanTD,
author = {Wang, Yujia and Zhang, Fang-Lue and Dodgson, Neil A.},
title = {ScanTD: 360° Scanpath Prediction based on Time-Series Diffusion},
year = {2024},
isbn = {9798400706868},
publisher = {Association for Computing Machinery},
address = {New York, NY, USA},
url = {https://doi.org/10.1145/3664647.3681315},
doi = {10.1145/3664647.3681315},
booktitle = {Proceedings of the 32nd ACM International Conference on Multimedia},
pages = {7764–7773},
numpages = {10},
location = {Melbourne VIC, Australia},
series = {MM '24}
}

@article{Wang_Zhang_Dodgson_2025, 
title={Target Scanpath-Guided 360-Degree Image Enhancement}, 
volume={39}, 
url={https://ojs.aaai.org/index.php/AAAI/article/view/32881}, 
DOI={10.1609/aaai.v39i8.32881}, 
number={8}, 
journal={Proceedings of the AAAI Conference on Artificial Intelligence}, 
author={Wang, Yujia and others.}, 
year={2025}, 
month={Apr.}, 
pages={8169-8177}
}

@inproceedings{liao2025convex,
  title={Convex Relaxation for Robust Vanishing Point Estimation in Manhattan World},
  author={Liao, Bangyan and Zhao, Zhenjun and Li, Haoang and Zhou, Yi and Zeng, Yingping and Li, Hao and Liu, Peidong},
  booktitle={Proceedings of the Computer Vision and Pattern Recognition Conference},
  pages={15823--15832},
  year={2025}
}

@inproceedings{zhao2024balf,
  title={Balf: Simple and efficient blur aware local feature detector},
  author={Zhao, Zhenjun},
  booktitle={Proceedings of the IEEE/CVF Winter Conference on Applications of Computer Vision},
  pages={3362--3372},
  year={2024}
}

@inproceedings{liu2025consistency,
  title={Consistency of local and global flatness for federated learning},
  author={Liu, Junkang and Shang, Fanhua and Tian, Yuxuan and Liu, Hongying and Liu, Yuanyuan},
  booktitle={Proceedings of the 33rd ACM International Conference on Multimedia},
  pages={3875--3883},
  year={2025}
}

@inproceedings{liu2024fedbcgd,
  title={Fedbcgd: Communication-efficient accelerated block coordinate gradient descent for federated learning},
  author={Liu, Junkang and Shang, Fanhua and Liu, Yuanyuan and Liu, Hongying and Li, Yuangang and Gong, YunXiang},
  booktitle={Proceedings of the 32nd ACM International Conference on Multimedia},
  pages={2955--2963},
  year={2024}
}

@inproceedings{liu2025improving,
  title={Improving Generalization in Federated Learning with Highly Heterogeneous Data via Momentum-Based Stochastic Controlled Weight Averaging},
  author={Liu, Junkang and Liu, Yuanyuan and Shang, Fanhua and Liu, Hongying and Liu, Jin and Feng, Wei},
  booktitle={Forty-second International Conference on Machine Learning},
  year={2025}
}

@inproceedings{zhao2023benchmark,
  title={Benchmark for Evaluating Initialization of Visual-Inertial Odometry},
  author={Zhao, Zhenjun and Chen, Ben M},
  booktitle={2023 42nd Chinese Control Conference (CCC)},
  pages={3935--3940},
  year={2023},
  organization={IEEE}
}

@ARTICLE{TCP,
  author={Jiang, Wenmiao and Zhang, Shunkai and You, Shengqi and Feng, Pengbin and Lu, Zhengyang},
  journal={IEEE Access}, 
  title={Traditional Chinese Painting Completion via Hierarchical Optimal Transport}, 
  year={2025},
  volume={13},
  number={},
  pages={170681-170692},
  doi={10.1109/ACCESS.2025.3615386}}

@INPROCEEDINGS{In-Pipe,
  author={Liu, Hao and Li, Xiang and Zhang, Xiang and Liu, Gang and Lu, Mingquan},
  booktitle={2025 IEEE International Conference on Robotics and Automation (ICRA)}, 
  title={In-Pipe Navigation Development Environment and a Smooth Path Planning Method on Pipeline Surface}, 
  year={2025},
  volume={},
  number={},
  pages={128084-128090},
  doi={10.1109/ICRA55743.2025.11128124}}

@inproceedings{liudifflow3d,
  title={DifFlow3D: Hierarchical Diffusion Models for Uncertainty-Aware 3D Scene Flow Estimation},
  author={Liu, Jiuming and Ye, Weicai and Wang, Guangming and Jiang, Chaokang and Pan, Lei and Han, Jinru and Liu, Zhe and Zhang, Guofeng and Wang, Hesheng},
  booktitle={IEEE transactions on pattern analysis and machine intelligence},
  year={2025},
}

@inproceedings{zhou2024lidarptq,
  title={LiDAR-PTQ: Post-Training Quantization for Point Cloud 3D Object Detection},
  author={Zhou, Sifan and Li, Liang and Zhang, Xinyu and Zhang, Bo and Bai, Shipeng and Sun, Miao and Zhao, Ziyu and Lu, Xiaobo and Chu, Xiangxiang},
  booktitle={The Twelfth International Conference on Learning Representations (ICLR)},
  year={2024},
}

@article{zhou2025neural,
  title={Neural-driven image editing},
  author={Zhou, Pengfei and Xia, Jie and Peng, Xiaopeng and Zhao, Wangbo and Ye, Zilong and Li, Zekai and Yang, Suorong and Pan, Jiadong and Chen, Yuanxiang and Wang, Ziqiao and others},
  journal={arXiv preprint arXiv:2507.05397},
  year={2025}
}

@inproceedings{zhou2025opening,
  title={OpenING: A Comprehensive Benchmark for Judging Open-ended Interleaved Image-Text Generation},
  author={Zhou, Pengfei and Peng, Xiaopeng and Song, Jiajun and Li, Chuanhao and Xu, Zhaopan and Yang, Yue and Guo, Ziyao and Zhang, Hao and Lin, Yuqi and He, Yefei and others},
  booktitle={Proceedings of the Computer Vision and Pattern Recognition Conference},
  pages={56--66},
  year={2025}
}

@inproceedings{wang2025point4bit,
title={Point4Bit: Post Training 4-bit Quantization for Point Cloud 3D Detection},
author={Jianyu Wang and Yu Wang and Shengjie Zhao and Sifan Zhou},
booktitle={The Thirty-ninth Annual Conference on Neural Information Processing Systems},
year={2025},
}

@inproceedings{gsq,
    title = "{GSQ}-Tuning: Group-Shared Exponents Integer in Fully Quantized Training for {LLM}s On-Device Fine-tuning",
    author = "Zhou, Sifan  and
      Wang, Shuo  and
      Yuan, Zhihang  and
      Shi, Mingjia  and
      Shang, Yuzhang  and
      Yang, Dawei",
    booktitle = "Findings of the Association for Computational Linguistics: ACL 2025",
    month = jul,
    year = "2025",
    address = "Vienna, Austria",
    publisher = "Association for Computational Linguistics",
    pages = "22971--22988",
    ISBN = "979-8-89176-256-5",
}

@inproceedings{focustrack,
  title={FocusTrack: One-Stage Focus-and-Suppress Framework for 3D Point Cloud
Object Tracking},
  author = {Zhou, Sifan and Nie, Jiahao and Zhao, Ziyu and Cao, Yichao and Lu, Xiaobo},
  year={2025},
isbn = {9798400720352},
publisher = {Association for Computing Machinery},
address = {New York, NY, USA},
doi = {10.1145/3746027.3754781},
booktitle={Proceedings of the 33rd ACM International Conference on Multimedia},
pages = {7366–7375},
numpages = {10},
location = {Dublin, Ireland},
}

@inproceedings{lu2024mace,
	title={Mace: Mass concept erasure in diffusion models},
	author={Lu, Shilin and Wang, Zilan and Li, Leyang and Liu, Yanzhu and Kong, Adams Wai-Kin},
	booktitle={Proceedings of the IEEE/CVF Conference on Computer Vision and Pattern Recognition},
	pages={6430--6440},
	year={2024}
}

@inproceedings{lu2023tf,
	title={Tf-icon: Diffusion-based training-free cross-domain image composition},
	author={Lu, Shilin and Liu, Yanzhu and Kong, Adams Wai-Kin},
	booktitle={Proceedings of the IEEE/CVF International Conference on Computer Vision},
	pages={2294--2305},
	year={2023}
}

@inproceedings{zhang2018shufflenet,
  title={Shufflenet: An extremely efficient convolutional neural network for mobile devices},
  author={Zhang, Xiangyu and Zhou, Xinyu and Lin, Mengxiao and Sun, Jian},
  booktitle={Proceedings of the IEEE conference on computer vision and pattern recognition},
  pages={6848--6856},
  year={2018}
}

@inproceedings{sc3d,
  title={Leveraging shape completion for 3d siamese tracking},
  author={Giancola, Silvio and Zarzar, Jesus and Ghanem, Bernard and Giancola, Silvio and Zarzar, Jesus},
  booktitle={Proceedings of the IEEE/CVF conference on computer vision and pattern recognition},
  pages={1359--1368},
  year={2019}
}

@inproceedings{p2b,
  title={P2b: Point-to-box network for 3d object tracking in point clouds},
  author={Qi, Haozhe and Feng, Chen and Cao, Zhiguo and Zhao, Feng and Xiao, Yang},
  booktitle={Proceedings of the IEEE/CVF conference on computer vision and pattern recognition},
  pages={6329--6338},
  year={2020}
}

@inproceedings{m2track,
  title={Beyond 3d siamese tracking: A motion-centric paradigm for 3d single object tracking in point clouds},
  author={Zheng, Chaoda and Yan, Xu and Zhang, Haiming and Wang, Baoyuan and Cheng, Shenghui and Cui, Shuguang and Li, Zhen},
  booktitle={Proceedings of the IEEE/CVF Conference on Computer Vision and Pattern Recognition},
  pages={8111--8120},
  year={2022}
}

@inproceedings{votenet,
  title={Deep hough voting for 3d object detection in point clouds},
  author={Qi, Charles R and Litany, Or and He, Kaiming and Guibas, Leonidas J},
  booktitle={proceedings of the IEEE/CVF International Conference on Computer Vision},
  pages={9277--9286},
  year={2019}
}

@inproceedings{glt,
  title={GLT-T: Global-Local Transformer Voting for 3D Single Object Tracking in Point Clouds},
  author={Nie, Jiahao and He, Zhiwei and Yang, Yuxiang and Gao, Mingyu and Zhang, Jing},
  booktitle={Proceedings of the AAAI Conference on Artificial Intelligence},
  pages={1957--1965},
  year={2023}
}

@inproceedings{mlvsnet,
  title={Mlvsnet: Multi-level voting siamese network for 3d visual tracking},
  author={Wang, Zhoutao and Xie, Qian and Lai, Yu-Kun and Wu, Jing and Long, Kun and Wang, Jun},
  booktitle={Proceedings of the IEEE/CVF International Conference on Computer Vision},
  pages={3101--3110},
  year={2021}
}

@inproceedings{bat,
  title={Box-aware feature enhancement for single object tracking on point clouds},
  author={Zheng, Chaoda and Yan, Xu and Gao, Jiantao and Zhao, Weibing and Zhang, Wei and Li, Zhen and Cui, Shuguang},
  booktitle={Proceedings of the IEEE/CVF International Conference on Computer Vision},
  pages={13199--13208},
  year={2021}
}

@article{v2b,
  title={3D Siamese voxel-to-BEV tracker for sparse point clouds},
  author={Hui, Le and Wang, Lingpeng and Cheng, Mingmei and Xie, Jin and Yang, Jian},
  journal={Advances in Neural Information Processing Systems},
  volume={34},
  pages={28714--28727},
  year={2021}
}

@article{lttr,
  title={3d object tracking with transformer},
  author={Cui, Yubo and Fang, Zheng and Shan, Jiayao and Gu, Zuoxu and Zhou, Sifan},
  journal={British Machine Vision Conference},
  pages={1445--1458},
  year={2021}
}

@article{cui2019point,
  title={Point siamese network for person tracking using 3D point clouds},
  author={Cui, Yubo and Fang, Zheng and Zhou, Sifan},
  journal={Sensors},
  volume={20},
  number={1},
  pages={143},
  year={2019},
  publisher={MDPI}
}

@InProceedings{ryoo2021tokenlearner,
  title={TokenLearner: Adaptive Space-Time Tokenization for Videos},
  author={Ryoo, Michael S. and Piergiovanni, AJ and Arnab, Anurag and Dehghani, Mostafa and Angelova, Anelia},
  booktitle={Advances in Neural Information Processing Systems (NeurIPS)},
  year={2021}
}

@inproceedings{tan2025xtrack,
  title={Xtrack: Multimodal training boosts rgb-x video object trackers},
  author={Tan, Yuedong and Wu, Zongwei and Fu, Yuqian and Zhou, Zhuyun and Sun, Guolei and Zamfir, Eduard and Ma, Chao and Paudel, Danda and Van Gool, Luc and Timofte, Radu},
  booktitle={Proceedings of the IEEE/CVF International Conference on Computer Vision},
  pages={5734--5744},
  year={2025}
}

@article{brodermann2025cafuser,
  title={Cafuser: Condition-aware multimodal fusion for robust semantic perception of driving scenes},
  author={Br{\"o}dermann, Tim and Sakaridis, Christos and Fu, Yuqian and Van Gool, Luc},
  journal={IEEE Robotics and Automation Letters},
  year={2025},
  publisher={IEEE}
}

@article{lowe1987three,
  title={Three-dimensional object recognition from single two-dimensional images},
  author={Lowe, David G},
  journal={Artificial intelligence},
  volume={31},
  number={3},
  pages={355--395},
  year={1987},
  publisher={Elsevier}
}

@article{hu2022lora,
  title={Lora: Low-rank adaptation of large language models.},
  author={Hu, Edward J and Shen, Yelong and Wallis, Phillip and Allen-Zhu, Zeyuan and Li, Yuanzhi and Wang, Shean and Wang, Lu and Chen, Weizhu and others},
  journal={ICLR},
  volume={1},
  number={2},
  pages={3},
  year={2022}
}

@article{pillartrack,
  title={PillarTrack: Boosting Pillar Representation for Transformer-based 3D Single Object Tracking on Point Clouds},
  author={Xu, Weisheng and Zhou, Sifan and Xiong, Jiaqi and Zhao, Ziyu and Yuan, Zhihang},
  journal={arXiv preprint arXiv:2404.07495},
  year={2024}
}

@inproceedings{harris1988combined,
  title={A combined corner and edge detector},
  author={Harris, Chris and Stephens, Mike and others},
  booktitle={Alvey vision conference},
  volume={15},
  number={50},
  pages={10--5244},
  year={1988},
  organization={Manchester, UK}
}

@inproceedings{swin,
  title={Swin transformer: Hierarchical vision transformer using shifted windows},
  author={Liu, Ze and Lin, Yutong and Cao, Yue and Hu, Han and Wei, Yixuan and Zhang, Zheng and Lin, Stephen and Guo, Baining},
  booktitle={ICCV},
  year={2021}
}

@article{zhou2023fastpillars,
  title={FastPillars: A Deployment-friendly Pillar-based 3D Detector},
  author={Zhou, Sifan and Zhang, Xinyu and Chu, Xiangxiang and Zhang, Bo and Zhao, Ziyu and Lu, Xiaobo},
  journal={IEEE Transactions on Circuits and Systems for Video Technology}, 
  year={2025},
  doi={10.1109/TCSVT.2025.3633725}
}

@inproceedings{ptt,
  title={Ptt: Point-track-transformer module for 3d single object tracking in point clouds},
  author={Shan, Jiayao and Zhou, Sifan and Fang, Zheng and Cui, Yubo},
  booktitle={2021 IEEE/RSJ International Conference on Intelligent Robots and Systems (IROS)},
  pages={1310--1316},
  year={2021},
  organization={IEEE}
}

@article{tao2023dudb,
  title={DUDB: deep unfolding-based dual-branch feature fusion network for pan-sharpening remote sensing images},
  author={Tao, Hailin and Li, Jinjiang and Hua, Zhen and Zhang, Fan},
  journal={IEEE Transactions on Geoscience and Remote Sensing},
  volume={62},
  pages={1--17},
  year={2023},
  publisher={IEEE}
}

@article{zhang2023multi,
  title={Multi-scale video super-resolution transformer with polynomial approximation},
  author={Zhang, Fan and Chen, Gongguan and Wang, Hua and Li, Jinjiang and Zhang, Caiming},
  journal={IEEE Transactions on Circuits and Systems for Video Technology},
  volume={33},
  number={9},
  pages={4496--4506},
  year={2023},
  publisher={IEEE}
}

@inproceedings{Wang2025OneImage,
  title     = {Do Vision Language Models Infer Human Intention Without Visual Perspective-Taking? Towards a Scalable ``One-Image-Probe-All'' Dataset},
  author    = {Wang, B. and Li, Y. and Zhou, Q. and Leong, H.Y. and Zhao, T. and Ye, L. and Deng, H. and Luo, D. and Vasconcelos, N.},
  booktitle = {Proceedings of the ICML 2025 Workshop on Assessing World Models},
  year      = {2025},
  url       = {https://openreview.net/forum?id=iekoq1rv80} 
}

@inproceedings{Qiu2025IntentVCNet,
  title     = {IntentVCNet: Bridging Spatio-Temporal Gaps for Intention-Oriented Controllable Video Captioning},
  author    = {Qiu, T. and Gao, J. and Li, J. and Leong, H.Y. and Zhang, L.},
  booktitle = {Proceedings of ACM Multimedia (MM) 2025},
  year      = {2025},
  note      = {Accepted, In Press},
  url       = {https://arxiv.org/abs/2507.18531},
  doi       = {10.48550/arXiv.2507.18531}
}

@article{ptt-journal,
  title={Real-time 3D single object tracking with transformer},
  author={Shan, Jiayao and Zhou, Sifan and Cui, Yubo and Fang, Zheng},
  journal={IEEE Transactions on Multimedia},
  volume={25},
  pages={2339--2353},
  year={2022},
  publisher={IEEE}
}

@inproceedings{pttr,
  title={Pttr: Relational 3d point cloud object tracking with transformer},
  author={Zhou, Changqing and Luo, Zhipeng and Luo, Yueru and Liu, Tianrui and Pan, Liang and Cai, Zhongang and Zhao, Haiyu and Lu, Shijian},
  booktitle={Proceedings of the IEEE/CVF Conference on Computer Vision and Pattern Recognition},
  pages={8531--8540},
  year={2022}
}

@inproceedings{stnet,
  title={3d siamese transformer network for single object tracking on point clouds},
  author={Hui, Le and Wang, Lingpeng and Tang, Linghua and Lan, Kaihao and Xie, Jin and Yang, Jian},
  booktitle={Computer Vision--ECCV 2022: 17th European Conference, Tel Aviv, Israel, October 23--27, 2022, Proceedings, Part II},
  pages={293--310},
  year={2022},
  organization={Springer}
}

@inproceedings{cmt,
  title={CMT: Context-Matching-Guided Transformer for 3D Tracking in Point Clouds},
  author={Guo, Zhiyang and Mao, Yunyao and Zhou, Wengang and Wang, Min and Li, Houqiang},
  booktitle={Computer Vision--ECCV 2022: 17th European Conference, Tel Aviv, Israel, October 23--27, 2022, Proceedings, Part XXII},
  pages={95--111},
  year={2022},
  organization={Springer}
}

@article{fan2025beyond,
  title={Beyond Frame-wise Tracking: A Trajectory-based Paradigm for Efficient Point Cloud Tracking},
  author={Fan, BaiChen and Zhou, Sifan and Li, Jian and Zhao, Shibo and Cao, Muqing and Wang, Qin},
  journal={arXiv preprint arXiv:2509.11453},
  year={2025}
}

@article{xumambaquant,
  title={MambaQuant: Quantizing the Mamba Family with Variance Aligned Rotation Methods},
  author={Xu, Zukang and Yue, Yuxuan and Hu, Xing and Yang, Dawei and Yuan, Zhihang and Jiang, Zixu and Chen, Zhixuan and Zhou, Sifan and others},
  journal={The Thirteenth International Conference on Learning Representations},
  year={2025}
}

@article{hu2025ostquant,
  title={OstQuant: Refining Large Language Model Quantization with Orthogonal and Scaling Transformations for Better Distribution Fitting},
  author={Hu, Xing and Cheng, Yuan and Yang, Dawei and Xu, Zukang and Yuan, Zhihang and Yu, Jiangyong and Xu, Chen and Jiang, Zhe and Zhou, Sifan},
  journal={The Thirteenth International Conference on Learning Representations},
  year={2025}
}

@inproceedings{osp2b,
  title={OSP2B: One-Stage Point-to-Box Network for 3D Siamese Tracking},
  author={Nie, Jiahao and He, Zhiwei and Yang, Yuxiang and Bao, Zhengyi and Gao, Mingyu and Zhang, Jing},
  booktitle={Proceedings of the Thirty-Second International Joint Conference on Artificial Intelligence},
  pages={1285--1293},
  year={2023}
}

@inproceedings{kitti,
  title={Are we ready for autonomous driving? the kitti vision benchmark suite},
  author={Geiger, Andreas and Lenz, Philip and Urtasun, Raquel},
  booktitle={2012 IEEE conference on computer vision and pattern recognition},
  pages={3354--3361},
  year={2012},
  organization={IEEE}
}

@inproceedings{nuScenes,
  title={nuscenes: A multimodal dataset for autonomous driving},
  author={Caesar, Holger and Bankiti, Varun and Lang, Alex H and Vora, Sourabh and Liong, Venice Erin and Xu, Qiang and Krishnan, Anush and Pan, Yu and Baldan, Giancarlo and Beijbom, Oscar},
  booktitle={Proceedings of the IEEE/CVF conference on computer vision and pattern recognition},
  pages={11621--11631},
  year={2020}
}

@inproceedings{otb2013,
  title={Online object tracking: A benchmark},
  author={Wu, Yi and Lim, Jongwoo and Yang, Ming-Hsuan},
  booktitle={Proceedings of the IEEE conference on computer vision and pattern recognition},
  pages={2411--2418},
  year={2013}
}

@article{3dsiamrpn,
  title={3d-siamrpn: An end-to-end learning method for real-time 3d single object tracking using raw point cloud},
  author={Fang, Zheng and Zhou, Sifan and Cui, Yubo and Scherer, Sebastian},
  journal={IEEE Sensors Journal},
  volume={21},
  number={4},
  pages={4995--5011},
  year={2020},
  publisher={IEEE}
}

@inproceedings{waymo,
  title={Scalability in perception for autonomous driving: Waymo open dataset},
  author={Sun, Pei and Kretzschmar, Henrik and Dotiwalla, Xerxes and Chouard, Aurelien and Patnaik, Vijaysai and Tsui, Paul and Guo, James and Zhou, Yin and Chai, Yuning and Caine, Benjamin and others},
  booktitle={Proceedings of the IEEE/CVF conference on computer vision and pattern recognition},
  pages={2446--2454},
  year={2020}
}

@InProceedings{cxtrack,
    author    = {Xu, Tian-Xing and Guo, Yuan-Chen and Lai, Yu-Kun and Zhang, Song-Hai},
    title     = {CXTrack: Improving 3D Point Cloud Tracking With Contextual Information},
    booktitle = {Proceedings of the IEEE/CVF Conference on Computer Vision and Pattern Recognition (CVPR)},
    month     = {June},
    year      = {2023},
    pages     = {1084-1093}
}

@inproceedings{cutrack,
title={Towards Category Unification of 3D Single Object Tracking on Point Clouds},
author={Jiahao Nie and Zhiwei He and Xudong Lv and Xueyi Zhou and Dong-Kyu Chae and Fei Xie},
booktitle={The Twelfth International Conference on Learning Representations},
year={2024}
}

@article{mbptrack,
  title={MBPTrack: Improving 3D Point Cloud Tracking with Memory Networks and Box Priors},
  author={Xu, Tian-Xing and Guo, Yuan-Chen and Lai, Yu-Kun and Zhang, Song-Hai},
  journal={arXiv preprint arXiv:2303.05071},
  year={2023}
}

@inproceedings{synctrack,
  title={Synchronize Feature Extracting and Matching: A Single Branch Framework for 3D Object Tracking},
  author={Ma, Teli and Wang, Mengmeng and Xiao, Jimin and Wu, Huifeng and Liu, Yong},
  booktitle={Proceedings of the IEEE/CVF International Conference on Computer Vision},
  pages={9953--9963},
  year={2023}
}

@inproceedings{corpnet,
  title={Correlation Pyramid Network for 3D Single Object Tracking},
  author={Wang, Mengmeng and Ma, Teli and Zuo, Xingxing and Lv, Jiajun and Liu, Yong},
  booktitle={Proceedings of the IEEE/CVF Conference on Computer Vision and Pattern Recognition},
  pages={3215--3224},
  year={2023}
}

@article{pttr++,
  title={Exploring point-bev fusion for 3d point cloud object tracking with transformer},
  author={Luo, Zhipeng and Zhou, Changqing and Pan, Liang and Zhang, Gongjie and Liu, Tianrui and Luo, Yueru and Zhao, Haiyu and Liu, Ziwei and Lu, Shijian},
  journal={IEEE Transactions on Pattern Analysis and Machine Intelligence},
  year={2024},
  publisher={IEEE}
}

@article{m2track++,
  title={An Effective Motion-Centric Paradigm for 3D Single Object Tracking in Point Clouds},
  author={Zheng, Chaoda and Yan, Xu and Zhang, Haiming and Wang, Baoyuan and Cheng, Shenghui and Cui, Shuguang and Li, Zhen},
  journal={IEEE Transactions on Pattern Analysis and Machine Intelligence},
  year={2023},
  publisher={IEEE}
}

@inproceedings{eco,
  title={Eco: Efficient convolution operators for tracking},
  author={Danelljan, Martin and Bhat, Goutam and Shahbaz Khan, Fahad and Felsberg, Michael},
  booktitle={Proceedings of the IEEE conference on computer vision and pattern recognition},
  pages={6638--6646},
  year={2017}
}

@inproceedings{atom,
  title={Atom: Accurate tracking by overlap maximization},
  author={Danelljan, Martin and Bhat, Goutam and Khan, Fahad Shahbaz and Felsberg, Michael},
  booktitle={Proceedings of the IEEE/CVF conference on computer vision and pattern recognition},
  pages={4660--4669},
  year={2019}
}

@inproceedings{center,
  title={Center-based 3d object detection and tracking},
  author={Yin, Tianwei and Zhou, Xingyi and Krahenbuhl, Philipp},
  booktitle={Proceedings of the IEEE/CVF conference on computer vision and pattern recognition},
  pages={11784--11793},
  year={2021}
}

@article{p2p,
  title={P2P: Part-to-Part Motion Cues Guide a Strong Tracking Framework for LiDAR Point Clouds},
  author={Nie, Jiahao and Xie, Fei and Zhou, Sifan and Zhou, Xueyi and Chae, Dong-Kyu and He, Zhiwei},
  journal={International Journal of Computer Vision},
  pages={1--17},
  year={2025},
  publisher={Springer}
}

@article{kristan2016novel,
  title={A novel performance evaluation methodology for single-target trackers},
  author={Kristan, Matej and Matas, Jiri and Leonardis, Ale{\v{s}} and Voj{\'\i}{\v{r}}, Tom{\'a}{\v{s}} and Pflugfelder, Roman and Fernandez, Gustavo and Nebehay, Georg and Porikli, Fatih and {\v{C}}ehovin, Luka},
  journal={IEEE transactions on pattern analysis and machine intelligence},
  volume={38},
  number={11},
  pages={2137--2155},
  year={2016},
  publisher={IEEE}
}

@inproceedings{robust,
  title={Robust 3D Tracking with Quality-Aware Shape Completion},
  author={Zhang, Jingwen and Zhou, Zikun and Lu, Guangming and Tian, Jiandong and Pei, Wenjie},
  booktitle={Proceedings of the AAAI Conference on Artificial Intelligence},
  volume={38},
  number={7},
  pages={7160--7168},
  year={2024}
}

@inproceedings{lidarsot,
  title={Model-free vehicle tracking and state estimation in point cloud sequences},
  author={Pang, Ziqi and Li, Zhichao and Wang, Naiyan},
  booktitle={2021 IEEE/RSJ International Conference on Intelligent Robots and Systems (IROS)},
  pages={8075--8082},
  year={2021},
  organization={IEEE}
}

@article{zhao2024ost,
  title={OST: Efficient One-stream Network for 3D Single Object Tracking in Point Clouds},
  author={Zhao, Xiantong and Han, Yinan and Tian, Shengjing and Liu, Jian and Liu, Xiuping},
  journal={IEEE Transactions on Multimedia},
  year={2024},
  publisher={IEEE}
}

@article{voxeltrack,
  title={VoxelTrack: Exploring Voxel Representation for 3D Point Cloud Object Tracking},
  author={Lu, Yuxuan and Nie, Jiahao and He, Zhiwei and Gu, Hongjie and Lv, Xudong},
  journal={arXiv preprint arXiv:2408.02263},
  year={2024}
}

@inproceedings{lighttrack,
  title={Lighttrack: Finding lightweight neural networks for object tracking via one-shot architecture search},
  author={Yan, Bin and Peng, Houwen and Wu, Kan and Wang, Dong and Fu, Jianlong and Lu, Huchuan},
  booktitle={Proceedings of the IEEE/CVF conference on computer vision and pattern recognition},
  pages={15180--15189},
  year={2021}
}

@inproceedings{fear,
  title={FEAR: Fast, efficient, accurate and robust visual tracker},
  author={Borsuk, Vasyl and Vei, Roman and Kupyn, Orest and Martyniuk, Tetiana and Krashenyi, Igor and Matas, Ji{\v{r}}i},
  booktitle={European conference on computer vision},
  pages={644--663},
  year={2022},
  organization={Springer}
}

@article{hit,
  title={Exploiting Lightweight Hierarchical ViT and Dynamic Framework for Efficient Visual Tracking},
  author={Kang, Ben and Chen, Xin and Zhao, Jie and Bo, Chunjuan and Wang, Dong and Lu, Huchuan},
  journal={arXiv preprint arXiv:2506.20381},
  year={2025}
}

@inproceedings{blatter2023efficient,
  title={Efficient visual tracking with exemplar transformers},
  author={Blatter, Philippe and Kanakis, Menelaos and Danelljan, Martin and Van Gool, Luc},
  booktitle={Proceedings of the IEEE/CVF Winter conference on applications of computer vision},
  pages={1571--1581},
  year={2023}
}

@article{zhu2025exploring,
  title={Exploring dynamic transformer for efficient object tracking},
  author={Zhu, Jiawen and Chen, Xin and Diao, Haiwen and Li, Shuai and He, Jun-Yan and Li, Chenyang and Luo, Bin and Wang, Dong and Lu, Huchuan},
  journal={IEEE Transactions on Neural Networks and Learning Systems},
  year={2025},
  publisher={IEEE}
}
